\pdfoutput=1

\documentclass[onefignum,onetabnum]{dclass}

\usepackage{lipsum}
\usepackage{amsfonts}
\usepackage{subcaption}
\usepackage{graphicx}
\usepackage{epstopdf}
\usepackage{algorithmic}
\usepackage{dsfont} 
\usepackage{adjustbox}
\usepackage{booktabs}

\makeatletter
\@ifundefined{AddToHook}{}{%
  \AddToHook{cmd/appendix/before}{\def\cref@section@alias{appendix}}}
\makeatother

\definecolor{red}{RGB}{255, 0, 0}
\definecolor{blue}{RGB}{51, 51, 179}
\definecolor{green}{RGB}{0, 128, 0}
\definecolor{gray}{RGB}{146, 146, 146}

\newcommand{\pp}[3]{
    \ifnum #3=1
        \frac{\partial #1}{\partial #2}
    \else
        \frac{\partial^{#3} #1}{\partial #2^{#3}}
    \fi
}
\newcommand{\dd}[3]{
    \ifnum #3=1
        \frac{d #1}{d #2}
    \else
        \frac{d^{#3} #1}{d #2^{#3}}
    \fi
}
\newcommand{\DD}[3]{
    \ifnum #3=1
        \frac{D #1}{D #2}
    \else
        \frac{D^{#3} #1}{D #2^{#3}}
    \fi
}

\newcommand{\iid}{\emph{i.i.d.~}}

\newcommand{\Kxxp}{K ( \mathbf{x}, \mathbf{x}' )}

\newcommand{\Kxz}{K ( \mathbf{x}, \mathbf{Z} )}

\newcommand{\Kzxp}{K ( \mathbf{Z}, \mathbf{x}' )}

\newcommand{\Kzz}{K ( \mathbf{Z}, \mathbf{Z} )}
\newcommand{\KzzHp}{ K ( \mathbf{Z}, \mathbf{Z} \, ; \boldsymbol{\psi} )}

\graphicspath{ {figures/} }
\ifpdf
  \DeclareGraphicsExtensions{.eps,.pdf,.png,.jpg}
\else
  \DeclareGraphicsExtensions{.eps}
\fi

\usepackage{enumitem}
\setlist[enumerate]{leftmargin=.5in}
\setlist[itemize]{leftmargin=.5in}

\newsiamremark{remark}{Remark}
\newsiamremark{hypothesis}{Hypothesis}
\crefname{hypothesis}{Hypothesis}{Hypotheses}
\newsiamthm{claim}{Claim}
\newsiamremark{fact}{Fact}
\crefname{fact}{Fact}{Facts}

\headers{Sequential Sparse Gaussian Process Quantile Regression}{H.~Nicolas and O.~Le Ma\^itre}

\title{Sequential Sparse Gaussian Process Quantile Regression}

\author{
Hugo Nicolas\thanks{Inria, Center for Applied Mathematics, \'Ecole polytechnique, Institut Polytechnique de Paris, Palaiseau, France (\email{hugo.nicolas@polytechnique.edu}).}
\and Olivier Le Ma\^itre\thanks{CNRS, Center for Applied Mathematics, \'Ecole polytechnique, Institut Polytechnique de Paris, Palaiseau, France (\email{olivier.le-maitre@polytechnique.edu}).}}

\usepackage{amsopn}

\ifpdf
\hypersetup{
  pdftitle={Sequential Sparse Gaussian Process Quantile Regression},
  pdfauthor={H. Nicolas and O. Le Ma\^itre}
}
\fi

\begin{document}

\maketitle


\begin{abstract}
Quantile regression aims to estimate the conditional quantiles of a response variable from observed data.
In a Bayesian setting, Gaussian process quantile regression provides uncertainty quantification but faces significant computational challenges due to the nonconjugacy of the asymmetric Laplace likelihood and the cost of posterior inference. 
We develop a sparse Gaussian process framework in which the quantile function is represented through a reduced set of inducing variables and posterior inference is performed using a Laplace approximation. 
A decomposition of the predictive uncertainty into conditional-prior and posterior-induced variance components is then exploited to drive two complementary adaptive mechanisms: inducing-input infilling and data acquisition. 
These mechanisms are combined within a sequential algorithm that allocates computational effort toward the dominant source of predictive uncertainty and adaptively controls model complexity.
Numerical experiments on benchmark problems demonstrate the accuracy of the Laplace approximation, the benefits of variance-based inducing-input placement, and the effectiveness of the proposed sequential enrichment strategy compared with predefined data-acquisition strategies.
The method is finally applied to ship seakeeping under forecast uncertainty, where it outperforms a replication-based counterpart from the literature under limited simulation budgets.
\end{abstract}




\section{Introduction} \label{sec:introduction}

Quantile regression aims to estimate the conditional quantiles of a response variable given a set of input variables.
It provides a more complete characterization of the conditional distribution than mean regression, and is particularly valuable in applications involving uncertainty quantification, risk assessment, and decision-making under uncertainty.

We consider applications that rely on potentially expensive stochastic simulators, whose output at a given input location is random.
This randomness typically arises from uncontrolled sources, such as environmental conditions or manufacturing deviations, so that only realizations of the response can be observed.
This setting precludes joint input-uncertainty modeling~\cite{cakmak2020,nguyen2021a} and accommodates high-dimensional or unparameterized uncertainty.

The conditional quantiles must therefore be inferred directly from these realizations.
The asymmetric loss function introduced by Koenker and Bassett~\cite{koenker1978} enables such inference, since its minimizer coincides with the desired conditional quantile.
Building on this formulation, a wide range of quantile regression functional forms have been proposed, including linear models~\cite{koenker1978,yu2001}, spline-based methods~\cite{koenker1994,thompson2010}, random forests~\cite{meinshausen2006}, neural networks~\cite{jantre2021}, kernel methods~\cite{takeuchi2006}, and Gaussian processes (GPs)~\cite{quadrianto2009,plumlee2014}.
In the present work, we focus on GP quantile regression.

GPs provide a flexible nonparametric framework for modeling unknown functions while naturally quantifying predictive uncertainty.
Within a Bayesian formulation, uncertainty is represented through a posterior distribution over the quantile function and can be propagated to predictions.
A common choice for Bayesian quantile regression is to combine a GP prior with an asymmetric Laplace likelihood, for which maximum likelihood estimation recovers the frequentist quantile regression solution.
This combination offers a principled way to learn conditional quantiles together with their associated uncertainty.

Two challenges nevertheless limit the practical deployment of Bayesian quantile regression with GPs.
First, posterior inference is analytically intractable because the asymmetric Laplace likelihood is not conjugate to the GP prior.
Second, quantile estimation can require many observations, particularly where the response density at the quantile is low, whereas exact GP inference scales cubically with the number of observations.
This motivates sparse inducing representations and, for expensive simulators, informative data acquisition.

Several approximations were proposed to address the posterior intractability.
Sampling-based methods, such as Markov chain Monte Carlo (MCMC), are flexible for quantile regression tasks~\cite{yu2001,kozumi2011} but can become computationally demanding for large datasets and complex models.
Deterministic alternatives instead approximate the posterior directly, through expectation propagation (EP), variational inference, or the Laplace approximation.
In the context of GP quantile regression, EP was considered by Boukouvalas et al.~\cite{boukouvalas2012}, while Abeywardana and Ramos~\cite{abeywardana2015} proposed a variational approximation based on the location-scale mixture representation of the asymmetric Laplace distribution.

To improve scalability, sparse GP models introduce a reduced set of latent inducing variables that act as a compact representation of the latent function.
Recent GP quantile regression frameworks, such as the one proposed by Picheny et al.~\cite{picheny2022}, have adopted sparse variational GP formulations.
Sparse variational GPs approximate the posterior distribution over the inducing variables by a Gaussian distribution whose mean and covariance are typically optimized jointly~\cite{snelson2005,titsias2009,hensman2013,leibfried2022}.
While effective, these approaches typically specify the number $M$ of inducing points \emph{a priori}, and the dimension of the optimization problem grows quadratically with $M$.

In this work, we develop a sequential sparse GP framework in which a decomposition of predictive uncertainty provides a unified basis for posterior inference, adaptive model enrichment, and data acquisition. 
The proposed methodology addresses two key challenges in sparse GP quantile regression: the efficient approximation of the posterior distribution over the inducing variables and the adaptive control of model complexity.

The first contribution is a sparse Bayesian quantile regression formulation based on a Laplace approximation of the posterior distribution over the inducing variables. 
Rather than optimizing both the mean vector and covariance matrix of an approximate posterior distribution, as is done in sparse variational GPs, we recast inference as the determination of a maximum \emph{a posteriori} (MAP) estimate and the inversion of the associated Hessian matrix. 
This yields a tractable optimization problem whose number of dominant optimization variables decreases from $\mathcal{O}(M^2)$ for variational formulations to $\mathcal{O}(M)$ in the proposed approach.

The second contribution is an adaptive inducing-input infilling strategy, as the quality of the inducing-point set is crucial for the performance of sparse GPs~\cite{burt2020,vakili2021,moss2023}.
The inducing inputs are treated as adaptive model components rather than fixed design variables. 
Inspired by Burt et al.~\cite{burt2020} and Ober et al.~\cite{ober2024}, new inducing inputs are introduced sequentially by maximizing the integrated reduction in conditional-prior variance, thereby determining both their locations and their number.

The third contribution is an adaptive data-acquisition strategy based on the posterior uncertainty induced by the inducing variables. 
Additional training data are acquired preferentially in regions where posterior uncertainty over the inducing variables contributes most strongly to predictive uncertainty.

These two enrichment mechanisms act on distinct components of the predictive uncertainty: inducing-input infilling reduces the conditional-prior variance, whereas data acquisition reduces the posterior-induced variance.
Finally, both enrichment mechanisms are combined into a unified sequential algorithm. 
A variance-based switching criterion determines whether computational effort should be allocated to improving the inducing representation or to acquiring additional training data. 
The resulting procedure adaptively balances model complexity and data availability while targeting the dominant source of predictive uncertainty.

The remainder of the paper is organized as follows.
\Cref{sec:bayesian_quantile_regression} introduces the sparse GP quantile regression formulation and the Laplace approximation.
\Cref{sec:sequential_enrichment} presents the inducing-input infilling strategy, the adaptive data-acquisition procedure, and the sequential enrichment algorithm.
Numerical experiments, including an application in maritime engineering, are reported in \cref{sec:numerical_experiments}.
\Cref{sec:conclusion} concludes the paper, and technical derivations are collected in the appendices.


\section{Bayesian quantile regression} \label{sec:bayesian_quantile_regression}


\subsection{Inference problem} \label{subsec:inference_problem}
Let $f : \mathcal{X} \times \varOmega \to \mathbb{R}$ be a measurable function, where the input space $\mathcal{X} \subset \mathbb{R}^D$ is compact and $( \varOmega, \mathcal{F}, \mathbb{P} )$ denotes the underlying probability space.
Define $y ( \mathbf{x} ) := f ( \mathbf{x}, \cdot )$ as the random output of the stochastic process $f$ at a deterministic input $\mathbf{x} \in \mathcal{X}$.
For a prescribed quantile level $\tau \in (0, 1)$, the $\tau$-quantile of the distribution of $y ( \mathbf{x} )$ is defined as $q_\tau ( \mathbf{x} ) = \inf \{ \nu \in \mathbb{R} : \mathbb{P} \, ( y ( \mathbf{x} ) \leq \nu ) \geq \tau \}$.

Consider a training dataset $\mathcal{D} = \{ ( \mathrm{y}_n, \mathbf{x}_n ) \}_{n=1}^N$ of $N$ independent realizations $\mathrm{y}_n \in \mathbb{R}$ of random variables $y ( \mathbf{x}_n )$ and their associated input locations $\mathbf{x}_n \in \mathcal{X}$.
Equivalently, $\mathrm{y}_n = f ( \mathbf{x}_n, \omega_n )$ for independent draws $\omega_1, \ldots, \omega_N$ from $( \varOmega, \mathcal{F}, \mathbb{P} )$.
Quantile regression aims to learn the unknown $\tau$-quantile function $q_\tau : \mathcal{X} \to \mathbb{R}$ from the training data.

In this work, we adopt a Bayesian perspective on the quantile regression problem.
Bayesian inference specifies a probabilistic model for the data-generating process and offers a principled framework for updating the model with new observations, according to Bayes' theorem.
The deviation $\varepsilon$ from the $\tau$-quantile at an input $\mathbf{x} \in \mathcal{X}$ is a random variable defined as
\begin{equation}
    \varepsilon ( \mathbf{x} )
    =
    y ( \mathbf{x} ) - q_\tau ( \mathbf{x} )
    .
    \label{eq:deviation_from_quantile}
\end{equation}

Following Yu and Moyeed~\cite{yu2001}, the deviation is assumed to follow an asymmetric Laplace distribution.
We consider the asymmetric Laplace density introduced by Yu and Zhang~\cite{yu2005} and defined as
\begin{equation}
    f_\varepsilon ( \nu \, ; \alpha )
    =
    \frac{ \tau (1 - \tau) }{ \alpha } \exp \! \left( -\frac{ \rho_\tau ( \nu ) }{ \alpha } \right)
    ,
    \label{eq:asymmetric_laplace_density}
\end{equation}
where $\alpha > 0$ is a scale parameter and $\rho_\tau$ denotes the quantile check function.
It is given by
\begin{equation}    
    \rho_\tau ( \nu )
    =
    \left( \tau - \mathds{1}_{ \{ \nu \leq 0 \} } \right) \nu
    ,
    \label{eq:check_function}
\end{equation}
where $\mathds{1}_{ \{ \nu \leq 0 \} }$ denotes the indicator function that equals $1$ if $\nu \leq 0$ and $0$ otherwise.
The check function is nonnegative.
The scale parameter $\alpha$ encodes the spread of the conditional distribution about its $\tau$-quantile.
We assume that it is constant across the input space $\mathcal{X}$.
This assumption can be relaxed by allowing the scale parameter to vary with the input, as done by Picheny et al.~\cite{picheny2022}, who placed a GP prior over it.
Under a conditional independence assumption, the observations $\mathbf{y} = [ \mathrm{y}_1, \ldots, \mathrm{y}_N ]^\top$ are independent given the input locations $\mathbf{X} = [ \mathbf{x}_1, \ldots, \mathbf{x}_N ]$ and the $\tau$-quantile function $q_\tau$.
Consequently, the likelihood $p ( \mathbf{y} \mid \mathbf{X}, q_\tau )$ of the observed data factorizes as
\begin{equation}
    p ( \mathbf{y} \mid \mathbf{X}, q_\tau )
    =
    \prod_{n=1}^{N} f_\varepsilon ( \mathrm{y}_n - q_\tau ( \mathbf{x}_n ) \, ; \alpha ) 
    =
    \left( \frac{ \tau (1 - \tau) }{ \alpha } \right)^{N} \exp \! \left( - \frac{1}{\alpha} \sum_{n=1}^{N} \rho_\tau ( \mathrm{y}_n - q_\tau ( \mathbf{x}_n ) ) \right)
    .
    \label{eq:quantile_likelihood}
\end{equation}

We model the latent $\tau$-quantile function as a realization of a GP.
A GP is a collection of random variables, any finite number of which have a joint normal distribution~\cite[Section~2.2]{rasmussen2006}.
It defines a distribution over functions, fully specified by its mean and covariance functions.
The covariance function, also known as the \textit{kernel}, encodes the assumptions about the smoothness and structure of the latent function.
Without loss of generality, we consider a zero-mean GP.
The prior distribution over the $\tau$-quantile function is then
\begin{equation}
    \pi ( q_\tau )
    =
    \mathcal{GP} \big( 0, \kappa ( \cdot, \cdot \, ; \boldsymbol{\psi} ) \big)
    ,
    \label{eq:quantile_prior}
\end{equation}
where $\kappa ( \cdot, \cdot \, ; \boldsymbol{\psi} )$ denotes the kernel with hyperparameters $\boldsymbol{\psi} \in \varPsi$.
In the following, we adopt the anisotropic Mat\'ern $5/2$ kernel, defined as
\begin{equation}
    \kappa ( \mathbf{x}, \mathbf{x}' \, ; \boldsymbol{\psi} ) 
    = 
    \sigma_{\text{s}}^2 \left( 1 + \sqrt{5} \, r ( \mathbf{x}, \mathbf{x}' ) + \frac{5}{3} r^2 ( \mathbf{x}, \mathbf{x}' ) \right) \exp \! \left( -\sqrt{5} \, r ( \mathbf{x}, \mathbf{x}' ) \right)
    ,
    \label{eq:matern_kernel}
\end{equation}
where $\sigma_{\text{s}}^2 \in \mathbb{R}_{> 0}$ denotes the signal variance, and
\begin{equation}
    r ( \mathbf{x}, \mathbf{x}' )
    =
    \sqrt{ ( \mathbf{x} - \mathbf{x}' )^{\top} \boldsymbol{\Lambda}^{-1} ( \mathbf{x} - \mathbf{x}' ) }
    \label{eq:anisotropic_euclidean_distance}
\end{equation}
is the anisotropic Euclidean distance, with $\boldsymbol{\Lambda} := \operatorname{diag} ( \ell_1^2, \ldots, \ell_D^2 )$, where $\ell_d \in \mathbb{R}_{>0}$ is the lengthscale of input dimension $d = 1, \ldots, D$.
The hyperparameter vector is thus given by $\boldsymbol{\psi} = [ \sigma_{\text{s}}^2, \ell_1, \ldots, \ell_D ]^{\top}$.
In the remainder of this work, we simplify the notation by omitting hyperparameter dependencies that are not essential to the discussion.

Bayes' theorem defines the posterior distribution over the $\tau$-quantile function conditional on the training data $\mathcal{D} = ( \mathbf{y}, \mathbf{X} )$ as
\begin{equation}
    p ( q_\tau \mid \mathbf{y}, \mathbf{X} )
    =
    \frac{ p ( \mathbf{y} \mid \mathbf{X}, q_\tau ) \, \pi ( q_\tau ) }{ p ( \mathbf{y} \mid \mathbf{X} ) }
    .
    \label{eq:quantile_posterior}
\end{equation}
The marginal likelihood $p ( \mathbf{y} \mid \mathbf{X} )$ ensures the normalization of the posterior.
It is defined as the expectation of the likelihood $p ( \mathbf{y} \mid \mathbf{X}, q_\tau )$ under the GP prior $\pi ( q_\tau )$.


\subsection{Sparse formulation} \label{subsec:sparse_formulation}
We introduce latent auxiliary variables that act as anchors for the regression.
Each auxiliary variable represents an evaluation of the latent $\tau$-quantile function $q_\tau$ at a location $\mathbf{z}_m \in \mathcal{X}$.
Denoting the set of these locations by $\mathcal{Z} = \{ \mathbf{z}_m \}_{m=1}^M$, we define these auxiliary variables as $\boldsymbol{u} = [ q_\tau (\mathbf{z}_1), \ldots, q_\tau (\mathbf{z}_M) ]^{\top}$.
In the remainder, we refer to the elements of $\boldsymbol{u}$ as \textit{inducing variables}, and the locations in $\mathcal{Z}$ as \textit{inducing inputs}.
Crucially, their number satisfies $M \ll N$, which ensures scalable inference for large training datasets.
The inducing inputs may be taken either as a subset of the training inputs $\mathbf{X}$ or as an entirely distinct set of locations.
Their selection is a key aspect of the proposed method, as we shall see in \cref{subsec:inducing_input_infilling}.
Until then, the inducing inputs in $\mathcal{Z}$ are assumed fixed and known.


\subsubsection{Prior distribution} \label{par:prior_distribution}
Denote by $\mathbf{u}$ a realization of the random vector $\boldsymbol{u}$.
Since the $\tau$-quantile function carries a GP prior, the inducing variables, conditional on the inducing inputs $\mathbf{Z} = [ \mathbf{z}_1, \ldots, \mathbf{z}_M ]$, follow a multivariate normal distribution.
Its probability density function (PDF) is
\begin{equation}
    \pi ( \mathbf{u} \mid \mathbf{Z} )
    =
    \mathcal{N} \big( \mathbf{u} \mid \boldsymbol{0}, \Kzz \big)
    =
    \frac{ 1 }{ (2 \pi)^{\frac{M}{2}} \left\lvert \Kzz \right\rvert^{\frac{1}{2}} } \exp \! \left( - \frac{1}{2} \mathbf{u}^{\top} \Kzz^{-1} \mathbf{u} \right)
    ,
    \label{eq:inducing_variables_prior}
\end{equation}
where $\left\lvert \cdot \right\rvert$ denotes the determinant operator, and $\Kzz$ is the covariance matrix evaluated at the inducing inputs, with entries $ [ \Kzz ]_{ij} = \kappa ( \mathbf{z}_i, \mathbf{z}_j \, ; \boldsymbol{\psi} )$ for $i, j = 1, \ldots, M$.
The prior over the $\tau$-quantile function, conditional on the inducing variables, is
\begin{equation}
    \pi ( q_\tau \mid \mathbf{u}, \mathbf{Z} )
    =
    \mathcal{GP} \big( \mu ( \cdot \, ; \mathbf{u}, \mathbf{Z} ), \Sigma_z^2 ( \cdot, \cdot \, ; \mathbf{Z} ) \big)
    .
    \label{eq:quantile_conditional_prior}
\end{equation}
The mean and covariance functions are, respectively, given by
\begin{align}
    \mu ( \mathbf{x} \, ; \mathbf{u}, \mathbf{Z} )
    & =
    \Kxz \Kzz^{-1} \mathbf{u}
    , 
    \label{eq:quantile_conditional_prior_mean}
    \\
    \Sigma_z^2 ( \mathbf{x}, \mathbf{x}' \, ; \mathbf{Z} )
    & =
    \Kxxp - \Kxz \Kzz^{-1} \Kzxp
    .
    \label{eq:quantile_conditional_prior_covariance}
\end{align}


\subsubsection{Posterior distribution} \label{par:posterior_distribution}
The joint posterior over the $\tau$-quantile function and the inducing variables factorizes exactly as $p ( q_\tau \mid \mathbf{u}, \mathbf{Z}, \mathbf{y}, \mathbf{X} ) \, p ( \mathbf{u} \mid \mathbf{Z}, \mathbf{y}, \mathbf{X} )$.
We approximate it by replacing the first factor with the conditional prior \eqref{eq:quantile_conditional_prior}, which yields
\begin{equation}
    p ( q_\tau, \mathbf{u} \mid \mathbf{Z}, \mathbf{y}, \mathbf{X} )
    \simeq
    \pi ( q_\tau \mid \mathbf{u}, \mathbf{Z} ) \, p ( \mathbf{u} \mid \mathbf{Z}, \mathbf{y}, \mathbf{X} )
    .
    \label{eq:quantile_joint_posterior}
\end{equation}
The second factor is the posterior over the inducing variables, conditional on the training data.
By Bayes' theorem, it decomposes as
\begin{equation}
    p ( \mathbf{u} \mid \mathbf{Z}, \mathbf{y}, \mathbf{X} )
    =
    \frac{ p ( \mathbf{y} \mid \mathbf{X}, \mathbf{u}, \mathbf{Z} ) \, \pi ( \mathbf{u} \mid \mathbf{Z} ) }{ p ( \mathbf{y} \mid \mathbf{X} ) }
    .
    \label{eq:inducing_variables_posterior}
\end{equation}
The prior $\pi ( \mathbf{u} \mid \mathbf{Z} )$ over the inducing variables is given in \cref{eq:inducing_variables_prior}.
The conditional likelihood
\begin{equation}
    p ( \mathbf{y} \mid \mathbf{X}, \mathbf{u}, \mathbf{Z} )
    =
    \mathbb{E}_{ \pi ( q_\tau \mid \mathbf{u}, \mathbf{Z} ) } \big[ p ( \mathbf{y} \mid \mathbf{X}, q_\tau ) \big]
    ,
    \label{eq:conditional_likelihood}
\end{equation}
marginalizes the data likelihood $p ( \mathbf{y} \mid \mathbf{X}, q_\tau )$, defined in \cref{eq:quantile_likelihood}, over the conditional prior \eqref{eq:quantile_conditional_prior}.

The sparse posterior quantile model follows by marginalizing \cref{eq:quantile_joint_posterior} over $\mathbf{u}$.
In classical GP regression, this step is analytical, since the posterior over the inducing variables is Gaussian.
Here, however, the training values consist of realizations of the response variable $y ( \mathbf{x} )$ rather than direct observations of the conditional $\tau$-quantile.
Formally, the asymmetric Laplace likelihood is not conjugate to the GP prior, rendering the posterior $p ( \mathbf{u} \mid \mathbf{Z}, \mathbf{y}, \mathbf{X} )$ over the inducing variables analytically intractable.
Approximating this posterior restores analytical tractability and is the subject of \cref{subsec:approximate_inference}.


\subsection{Approximate inference} \label{subsec:approximate_inference}


\subsubsection{Laplace approximation} \label{subsubsec:laplace_approximation}
We approximate the posterior $p ( \mathbf{u} \mid \mathbf{Z}, \mathbf{y}, \mathbf{X} )$ over the inducing variables by the Laplace approximation, which replaces a density with a Gaussian centered at its mode~\cite[Section~4.6.8]{murphy2022}.
This mode is the MAP estimate $\hat{\mathbf{u}}$ of the inducing variables.
Since the denominator in \cref{eq:inducing_variables_posterior} does not depend on $\mathbf{u}$, $\hat{\mathbf{u}}$ also maximizes the logarithm of the joint density $p ( \mathbf{u}, \mathbf{y} \mid \mathbf{X}, \mathbf{Z} )$,
\begin{equation}
    \hat{\mathbf{u}}
    =
    \arg \max_{ \mathbf{u} \in \mathbb{R}^M } \mathcal{J}_N ( \mathbf{u} )
    , \quad \text{with} \quad
    \mathcal{J}_N ( \mathbf{u} )
    :=
    \log p ( \mathbf{u}, \mathbf{y} \mid \mathbf{X}, \mathbf{Z})
    =
    \log p ( \mathbf{y} \mid \mathbf{X}, \mathbf{u}, \mathbf{Z} ) + \log \pi ( \mathbf{u} \mid \mathbf{Z} )
    .
    \label{eq:map_estimation}
\end{equation}

The conditional likelihood $p ( \mathbf{y} \mid \mathbf{X}, \mathbf{u}, \mathbf{Z} )$, given in \cref{eq:conditional_likelihood}, does not admit a closed-form expression and would therefore require approximation by sampling.
Repeated sampling would be prohibitively expensive given the sequential refinement strategy proposed in the present work.
To overcome this limitation, we propose to use a surrogate $\widetilde{\mathcal{J}}_N$ of the log-joint density $\mathcal{J}_N ( \mathbf{u} ) = \log ( \mathbb{E}_{ \pi ( q_\tau \mid \mathbf{u}, \mathbf{Z} ) } \big[ p ( \mathbf{y} \mid \mathbf{X}, q_\tau ) \big] ) + \log \pi ( \mathbf{u} \mid \mathbf{Z} )$.
The surrogate is obtained by interchanging the expectation and the logarithm, yielding
\begin{equation}
    \widetilde{\mathcal{J}}_N ( \mathbf{u} )
    :=
    \mathbb{E}_{ \pi ( q_\tau \mid \mathbf{u}, \mathbf{Z} ) } \big[ \log p ( \mathbf{y} \mid \mathbf{X}, q_\tau ) \big]
    +
    \log \pi ( \mathbf{u} \mid \mathbf{Z} )
    .
    \label{eq:log_joint_surrogate}
\end{equation}
By Jensen's inequality, $\mathbb{E}_{ \pi ( q_\tau \mid \mathbf{u}, \mathbf{Z} ) } \big[ \log p ( \mathbf{y} \mid \mathbf{X}, q_\tau ) \big] \le \log \! \left( \mathbb{E}_{ \pi ( q_\tau \mid \mathbf{u}, \mathbf{Z} ) } \big[ p ( \mathbf{y} \mid \mathbf{X}, q_\tau ) \big]  \right)$, so this surrogate is a lower bound on the exact log-joint density.
Maximizing it amounts to maximizing a tractable, conservative approximation.
\Cref{app:control_of_the_lower_bound_surrogate_error} bounds the gap between the two objectives in terms of the joint conditional-prior covariance at the training inputs, connecting the accuracy of the surrogate to the inducing-input representation.
Although the effect of the surrogate on the location of the MAP estimate $\hat{\mathbf{u}}$ is not analyzed theoretically in this work, the numerical results support its practical effectiveness.
The surrogate moreover has a closed-form expression, in which Gaussian convolution smooths out the nondifferentiability of the check function $\rho_\tau$, provided that the conditional-prior variance $\Sigma_z^2 ( \mathbf{x}_n, \mathbf{x}_n \, ; \mathbf{Z} )$ is positive at every training input.
This variance vanishes only at a training input that coincides with an inducing input, and adding a small jitter to $\Kzz$ regularizes that case.
Combined with the $C^{\infty}$ smoothness of the Gaussian prior in $\mathbf{u}$ and the linearity in $\mathbf{u}$ of the conditional-prior mean function $\mu$, the Gaussian convolution renders $\widetilde{\mathcal{J}}_N ( \mathbf{u} )$ of class $C^{\infty}$.
Its gradient and Hessian are available analytically, the latter being given in \cref{app:analytical_expressions}.
These characteristics enable solving \cref{eq:map_estimation}, with $\widetilde{\mathcal J}_N$ in place of $\mathcal{J}_N$, by efficient deterministic gradient-based optimization algorithms.
The availability of the Hessian can be further exploited to accelerate their convergence.
For these reasons, the surrogate $\widetilde{\mathcal{J}}_N$ is used in the rest of the derivation, and $\hat{\mathbf{u}}$ hereafter denotes its maximizer.

The surrogate posterior $\widetilde{p} ( \mathbf{u} \mid \mathbf{Z}, \mathbf{y}, \mathbf{X} ) \propto \exp ( \widetilde{\mathcal{J}}_N ( \mathbf{u} ) )$ is approximated by a Gaussian through a second-order Taylor expansion of its logarithm about $\hat{\mathbf{u}}$,
\begin{equation}
    \widetilde{p} ( \mathbf{u} \mid \mathbf{Z}, \mathbf{y}, \mathbf{X} )
    \simeq
    \hat{p} ( \mathbf{u} \mid \mathbf{Z}, \mathbf{y}, \mathbf{X} )
    :=
    \mathcal{N} \big( \mathbf{u} \mid \hat{\mathbf{u}}, \hat{\mathbf{C}} \big)
    ,
    \label{eq:inducing_variables_posterior_gaussian_approximation}
\end{equation}
where $\hat{\mathbf{C}}$ denotes the inverse of the negative Hessian of $\widetilde{\mathcal{J}}_N$ at its maximizer:
\begin{equation}
    \hat{\mathbf{C}}
    =
    [ - \mathbf{H}_N ( \hat{\mathbf{u}} ) ]^{-1}
    ,
    \quad \text{with} \quad
    \mathbf{H}_N ( \hat{\mathbf{u}} )
    :=
    \nabla_\mathbf{u}^2 \, \widetilde{\mathcal{J}}_N ( \mathbf{u} ) \vert_{ \mathbf{u} = \hat{\mathbf{u}} }
    .
    \label{eq:laplace_approximation_covariance}
\end{equation}

In practice, the GP hyperparameters $\boldsymbol{\psi}$ and the asymmetric Laplace scale parameter $\alpha$ enter the MAP objective.
We therefore reintroduce them explicitly into the notation and optimize them alongside $\mathbf{u}$, leading to the bilevel optimization problem
\begin{subequations} \label{eq:bilevel_optimization_problem}
    \begin{align}
        \big( \hat{\boldsymbol{\psi}}, \hat{\alpha} \big)
        & =
        \mathop{\arg \max}_{ ( \boldsymbol{\psi}, \alpha ) \in \varPsi \times \mathbb{R}_{>0} }
        \hat{\mathcal{F}}_N ( \boldsymbol{\psi}, \alpha )
        ,
        \label{eq:upper_level_optimization_problem}
        \\
        \text{subject to} \quad
        \hat{\mathbf{u}} ( \boldsymbol{\psi}, \alpha )
        & =
        \arg \max_{ \mathbf{u} \in \mathbb{R}^M }
        \widetilde{\mathcal{J}}_N ( \mathbf{u} \, ; \boldsymbol{\psi}, \alpha )
        ,
        \label{eq:lower_level_optimization_problem}
    \end{align}
\end{subequations}
where $\hat{\mathcal{F}}_N$ approximates the log-marginal likelihood $\mathcal{F}_N ( \boldsymbol{\psi}, \alpha ) := \log \int \exp ( \mathcal{J}_N ( \mathbf{u} \, ; \boldsymbol{\psi}, \alpha ) ) \, d \mathbf{u}$.
Replacing $\mathcal{J}_N$ with $\widetilde{\mathcal{J}}_N$ in the integrand and applying the Laplace approximation yields
\begin{subequations}
    \begin{align}
        \widetilde{\mathcal{F}}_N ( \boldsymbol{\psi}, \alpha )
        & :=
        \log \int_{\mathbb{R}^M} \exp \! \left( \widetilde{\mathcal{J}}_N ( \mathbf{u} \, ; \boldsymbol{\psi}, \alpha ) \right) \, d \mathbf{u}
        \\
        & \simeq
        \hat{\mathcal{F}}_N ( \boldsymbol{\psi}, \alpha )
        :=
        \widetilde{\mathcal{J}}_N ( \hat{\mathbf{u}} ( \boldsymbol{\psi}, \alpha ) \, ; \boldsymbol{\psi}, \alpha ) - \frac{1}{2} \log \left\lvert - \mathbf{H}_N ( \hat{\mathbf{u}} ( \boldsymbol{\psi}, \alpha ) \, ; \boldsymbol{\psi}, \alpha ) \right\rvert + \frac{M}{2} \log ( 2 \pi )
        .
        \label{eq:log_marginal_likelihood_approximation}
    \end{align}
\end{subequations}
To improve numerical conditioning, the inner problem \cref{eq:lower_level_optimization_problem} is solved in the whitened parameterization $\mathbf{u} = \mathbf{L} \mathbf{v}$~\cite{murray2010}, where $\mathbf{L}$ is the lower Cholesky factor of $\Kzz$, so that $\mathbf{v} \sim \mathcal{N} ( \mathbf{0}, \mathbf{I}_M )$ under the prior.


\subsubsection{Predictive quantile distribution} \label{subsubsec:predictive_distribution}
The MAP estimate $\hat{\mathbf{u}}$ and the covariance matrix $\hat{\mathbf{C}}$ in the Laplace approximation are obtained for a specific training dataset $\mathcal{D} = ( \mathbf{y}, \mathbf{X} )$.
We make this dependence explicit by writing these quantities as $\hat{\mathbf{u}} ( \mathbf{y}, \mathbf{X} )$ and $\hat{\mathbf{C}} ( \mathbf{y}, \mathbf{X} )$, respectively.
Substituting the Gaussian approximation \eqref{eq:inducing_variables_posterior_gaussian_approximation} into the sparse approximation \eqref{eq:quantile_joint_posterior} and marginalizing over $\mathbf{u}$ yields the approximate posterior predictive distribution over the $\tau$-quantile function
\begin{equation}
    \hat{p} ( q_\tau \mid \mathbf{y}, \mathbf{X}, \mathbf{Z} )
    =
    \mathbb{E}_{ \hat{p} ( \mathbf{u} \mid \mathbf{Z}, \mathbf{y}, \mathbf{X} ) } \big[ \pi ( q_\tau \mid \mathbf{u}, \mathbf{Z} ) \big]                
    =
    \mathcal{GP} \big( \hat{q}_\tau ( \cdot \, ; \mathbf{y}, \mathbf{X}, \mathbf{Z} ), \Sigma_z^2 ( \cdot, \cdot \, ; \mathbf{Z} ) + \Sigma_u^2 ( \cdot, \cdot \, ; \mathbf{y}, \mathbf{X}, \mathbf{Z} ) \big)
    ,
    \label{eq:quantile_predictive_distribution}
\end{equation}
where $\hat{q}_\tau$ denotes the posterior predictive mean function, and the posterior predictive covariance function decomposes into the conditional-prior covariance $\Sigma_z^2$ and the posterior-induced covariance $\Sigma_u^2$.
Their explicit expressions are given by
\begin{align}
    \hat{q}_\tau ( \mathbf{x} \, ; \mathbf{y}, \mathbf{X}, \mathbf{Z} )
    & =
    \Kxz \Kzz^{-1} \hat{\mathbf{u}} ( \mathbf{y}, \mathbf{X} )
    , 
    \label{eq:quantile_predictive_distribution_mean}
    \\
    \Sigma_z^2 ( \mathbf{x}, \mathbf{x}' \, ; \mathbf{Z} )
    & =
    \Kxxp - \Kxz \Kzz^{-1} \Kzxp
    , 
    \label{eq:quantile_predictive_distribution_conditional_prior_covariance}
    \\
    \Sigma^2_u ( \mathbf{x}, \mathbf{x}' \, ; \mathbf{y}, \mathbf{X}, \mathbf{Z} )
    & =
    \Kxz \Kzz^{-1} \hat{\mathbf{C}} ( \mathbf{y}, \mathbf{X} ) \Kzz^{-1} \Kzxp
    .
    \label{eq:quantile_predictive_distribution_posterior_induced_covariance}
\end{align}


\section{Sequential enrichment} \label{sec:sequential_enrichment}

We adopt the integrated mean squared error (IMSE)~\cite{sacks1989} as a performance metric to quantify predictive accuracy.
For the present quantile model, it is expressed as
\begin{subequations} \label{eq:imse}
    \begin{align}
        \text{IMSE}
        & =
        \int_{\mathcal{X}} \mathbb{E}_{ \hat{p} ( q_\tau \mid \mathbf{y}, \mathbf{X}, \mathbf{Z} ) } \big[ \left( q_\tau ( \mathbf{x} ) - q_{\tau, \text{true}} ( \mathbf{x} ) \right)^2 \big] \, d \mathbf{x}
        \\
        & =
        \int_{\mathcal{X}} \Big\{ \underbrace{ \Sigma_z^2 ( \mathbf{x}, \mathbf{x} \, ; \mathbf{Z} ) + \Sigma_u^2 ( \mathbf{x}, \mathbf{x} \, ; \mathbf{y}, \mathbf{X}, \mathbf{Z} ) }_{ \text{variance} } + \underbrace{ \left( \hat{q}_\tau ( \mathbf{x} \, ; \mathbf{y}, \mathbf{X}, \mathbf{Z} ) - q_{\tau, \text{true}} ( \mathbf{x} ) \right)^2 }_{ \text{best-prediction squared error} } \Big\} \, d \mathbf{x}
        ,
        \label{eq:imse_developed_form}
    \end{align}
\end{subequations}
where $q_{\tau, \text{true}} ( \mathbf{x} )$ denotes the true $\tau$-quantile of the distribution of $y ( \mathbf{x} )$.
Uniform weighting is used in \cref{eq:imse} and retained throughout.
A prescribed continuous input distribution can be mapped to independent uniform variables through an isoprobabilistic transformation, so that uniform integration in the transformed coordinates is equivalent to distribution-weighted integration in the physical coordinates.

The first variance term in \cref{eq:imse_developed_form} is the conditional-prior variance evaluated at input $\mathbf{x} \in \mathcal{X}$.
With fixed GP hyperparameters, it is reduced by adding distinct inducing inputs.
Judicious inducing-input placement accelerates this reduction.
The second variance term in \cref{eq:imse_developed_form} reflects the variability in the quantile predictions induced by the posterior uncertainty over the inducing variables $\boldsymbol{u}$.
This component is reduced by improving the estimation of the inducing variables.
With fixed hyperparameters, this can be achieved by augmenting the training dataset $\mathcal{D} = ( \mathbf{y}, \mathbf{X} )$ with additional observations.
Both variance terms can be computed in closed form, respectively through the covariance functions \eqref{eq:quantile_predictive_distribution_conditional_prior_covariance} and \eqref{eq:quantile_predictive_distribution_posterior_induced_covariance}.
The final term in \cref{eq:imse_developed_form}, referred to as the best-prediction squared error, measures the squared deviation between the posterior predictive mean and the true $\tau$-quantile.
It is typically not available since the true $\tau$-quantile is unknown.

We present a sequential enrichment strategy to improve the global accuracy of the posterior quantile model.
The strategy comprises two adaptive components, each targeting a distinct variance contribution in the IMSE \eqref{eq:imse_developed_form}.
The first component is an inducing-input infilling procedure that maximizes the integrated reduction in conditional-prior variance.
This yields a systematic placement of inducing inputs $\mathbf{Z}$ for a given training dataset $\mathcal{D}$.
The second component concerns the acquisition of new observations of the stochastic process $y ( \cdot )$.
It determines where training data should be acquired, based on the posterior-induced variance of the posterior quantile model, whose complexity is governed by the number and placement of the inducing inputs $\mathbf{Z}$.
The two adaptive components are repeated until convergence or until the computational budget is exhausted.
A criterion based on the dominant variance contribution is proposed to switch between the two components at each iteration of the sequential algorithm.
This provides a principled framework for sequentially targeting the most prominent source of global predictive error across the input space $\mathcal{X}$.

The infilling of inducing inputs is discussed in \cref{subsec:inducing_input_infilling}.
The procedure for acquiring new observations of the data-generating process is detailed in \cref{subsec:training_data_acquisition}.
The switching criterion and the complete algorithm are presented in \cref{subsec:sequential_algorithm}.
Computational considerations are discussed in \cref{subsec:computational_considerations}.


\subsection{Inducing-input infilling} \label{subsec:inducing_input_infilling}
In this section, we treat the training dataset $\mathcal{D}$ as fixed and address the infilling of the inducing-input set $\mathcal{Z}$.

With fixed GP hyperparameters, adding new inducing inputs reduces the conditional-prior variance.
Assuming that this addition does not significantly alter the remaining terms in \cref{eq:imse_developed_form}, global predictive accuracy can be enhanced by targeting the conditional-prior variance contribution.
We thus propose to add inducing inputs to $\mathcal{Z}$ at locations that maximize the integrated reduction in conditional-prior variance over the input space $\mathcal{X}$.
This strategy is expected to yield performance comparable to, or exceeding that of, greedy strategies based on the maximum conditional-prior variance, particularly in high-dimensional settings.
See~\cite{sacks1989,gramacy2009,bect2012} for related work on optimal experimental design with standard GP regression.

The conditional-prior variance at an arbitrary input $\mathbf{x} \in \mathcal{X}$, given the current inducing inputs $\mathbf{Z}$ and a candidate inducing input $\tilde{\mathbf{z}} \in \mathcal{X}$, is expressed as
\begin{equation}
    \Sigma_z^2 ( \mathbf{x}, \mathbf{x} \, ; [ \mathbf{Z}, \tilde{\mathbf{z}} ] )
    =
    K ( \mathbf{x}, \mathbf{x} )
    -
    \begin{bmatrix}
        K ( \mathbf{Z}, \mathbf{x} ) \\
        K ( \tilde{\mathbf{z}}, \mathbf{x} )
    \end{bmatrix}^{\top}
    \begin{bmatrix}
        \Kzz & K ( \tilde{\mathbf{z}}, \mathbf{Z} )^{\top} \\
        K ( \tilde{\mathbf{z}}, \mathbf{Z} ) & K ( \tilde{\mathbf{z}}, \tilde{\mathbf{z}} )
    \end{bmatrix}^{-1}
    \begin{bmatrix}
        K ( \mathbf{Z}, \mathbf{x} ) \\
        K ( \tilde{\mathbf{z}}, \mathbf{x} )
    \end{bmatrix}
    \label{eq:quantile_conditional_prior_variance_with_new_inducing_input}
    .
\end{equation}
To avoid direct inversion of the block matrix, we apply the block matrix inversion identity based on the Schur complement, expressing the inverse in terms of $\Kzz^{-1}$.
Since $\Kzz^{-1}$ is already available from the training of the quantile model, \cref{eq:quantile_conditional_prior_variance_with_new_inducing_input} becomes significantly cheaper to evaluate.
The integrated reduction in conditional-prior variance induced by adding $\tilde{\mathbf{z}}$ is then
\begin{subequations} \label{eq:integrated_variance_reduction}
    \begin{align}
        \Delta_z ( \tilde{\mathbf{z}} \, ; \mathbf{Z} )
        & :=
        \int_{\mathcal{X}} \Big\{ \Sigma_z^2 ( \mathbf{x}, \mathbf{x} \, ; \mathbf{Z} ) - \Sigma_z^2 ( \mathbf{x}, \mathbf{x} \, ; [ \mathbf{Z}, \tilde{\mathbf{z}} ] ) \Big\} \, d \mathbf{x}
        \label{eq:integrated_conditional_prior_variance_reduction_definition}
        \\
        & =
        \int_{\mathcal{X}} \Bigg\{ \frac{ \left( K ( \tilde{\mathbf{z}}, \mathbf{x} ) - K ( \tilde{\mathbf{z}}, \mathbf{Z} ) \Kzz^{-1} K ( \mathbf{Z}, \mathbf{x} ) \right)^2  }{ K ( \tilde{\mathbf{z}}, \tilde{\mathbf{z}} ) - K ( \tilde{\mathbf{z}}, \mathbf{Z} ) \Kzz^{-1} K ( \tilde{\mathbf{z}}, \mathbf{Z} )^{\top} } \Bigg\} \, d \mathbf{x}
        .
        \label{eq:integrated_conditional_prior_variance_reduction_developed_form}
    \end{align}
\end{subequations}
The integral over $\mathcal{X}$ can be estimated through Monte Carlo sampling.

Throughout the infilling procedure, each new inducing input $\mathbf{z}_{\text{new}} \in \mathcal{X}$ is selected as the maximizer of the integrated reduction in conditional-prior variance \eqref{eq:integrated_variance_reduction}:
\begin{equation}
    \mathbf{z}_{\text{new}}
    =
    \arg \max_{ \tilde{\mathbf{z}} \in \mathcal{X} } \Delta_z ( \tilde{\mathbf{z}} \, ; \mathbf{Z} )
    .
    \label{eq:optimal_inducing_input_location}
\end{equation}
The gradient of \cref{eq:integrated_conditional_prior_variance_reduction_developed_form} with respect to $\tilde{\mathbf{z}}$ is available in closed form.
Consequently, the optimization problem \eqref{eq:optimal_inducing_input_location} can be solved with a deterministic, gradient-based algorithm.


\subsection{Training-data acquisition} \label{subsec:training_data_acquisition}
We now address the problem of augmenting the training dataset $\mathcal{D}$ with additional observations of the stochastic process $y ( \cdot )$, while keeping the inducing-input set $\mathcal{Z}$ fixed.

Since the inducing inputs $\mathbf{Z}$ remain unchanged, enriching the training dataset does not significantly alter the conditional-prior variance.
We thus propose to acquire observations at locations selected via rejection sampling, with acceptance probabilities governed by the posterior-induced variance.
This adaptive strategy is expected to improve training-data acquisition efficiency by focusing on regions where posterior uncertainty over the inducing variables induces high variability in the quantile predictions.

The rejection sampling algorithm is based on an upper bound $\sigma_{\text{ref}}^2$ for the posterior-induced variance:
\begin{equation}
    \sigma_{\text{ref}}^2 \ge \max_{ \mathbf{x} \in \mathcal{X} } \Sigma_u^2 ( \mathbf{x}, \mathbf{x} \, ; \mathbf{y}, \mathbf{X}, \mathbf{Z} ).
    \label{eq:rejection_sampling_reference_input}
\end{equation}
This upper bound serves as a reference value for the acceptance probabilities.
At each iteration of the rejection sampling, a candidate input location $\mathbf{x}_{\text{prop}}$ is drawn uniformly over the input space $\mathcal{X}$ and accepted if
\begin{equation}
    \upsilon < f_{\text{accept}} \! \left( \frac{ \Sigma_u^2 ( \mathbf{x}_{\text{prop}}, \mathbf{x}_{\text{prop}} \, ; \mathbf{y}, \mathbf{X}, \mathbf{Z} ) }{ \sigma_{\text{ref}}^2 } \right)
    ,
    \label{eq:rejection_sampling_acceptance}
\end{equation}
where $\upsilon$ is a random variable uniformly distributed in $( 0, 1 )$, and $f_{\text{accept}} : [0, 1] \to [0, 1]$ is a monotone acceptance function. 
In this work, we use the identity function $f_\text{accept} ( x ) = x$.
Alternative acceptance functions can be used to adjust the selectivity of the sampling process.
Since the gradient of \cref{eq:quantile_predictive_distribution_posterior_induced_covariance} with respect to $\mathbf{x}$ is available in closed form, we set the upper bound to the maximum of $\Sigma_u^2 ( \mathbf{x}, \mathbf{x} \, ; \mathbf{y}, \mathbf{X}, \mathbf{Z} )$, found with a deterministic, gradient-based algorithm.


\subsection{Sequential algorithm} \label{subsec:sequential_algorithm}
\Cref{subsec:inducing_input_infilling} addressed the inducing-input infilling given a fixed training dataset $\mathcal{D}$.
\Cref{subsec:training_data_acquisition} addressed the adaptive acquisition of new training data given a fixed inducing-input set $\mathcal{Z}$, and thus a fixed structural complexity of the posterior quantile model.
The present section combines them into a sequential algorithm that alternates between the two according to a switching criterion.


\subsubsection{Switching criterion} \label{subsubsec:switching_criterion}
At each iteration, the sequential algorithm selects between adaptive training-data acquisition and inducing-input infilling based on the dominant variance contribution to the global error, as measured by the IMSE \eqref{eq:imse}.
A practical heuristic in this setting is to acquire new training data as long as the integrated posterior-induced variance $\Sigma_u^2 ( \mathbf{x}, \mathbf{x} \, ; \mathbf{y}, \mathbf{X}, \mathbf{Z} )$, evaluated over the input space $\mathcal{X}$, dominates the integrated conditional-prior variance $\Sigma_z^2 ( \mathbf{x}, \mathbf{x} \, ; \mathbf{Z} )$, evaluated over the same domain.
Inducing-input infilling is triggered once the reverse holds, formally when
\begin{equation}
    \int_{\mathcal{X}} \Sigma_z^2 ( \mathbf{x}, \mathbf{x} \, ; \mathbf{Z} ) \, d \mathbf{x}
    \geq c_{\text{ratio}}
    \int_{\mathcal{X}} \Sigma_u^2 ( \mathbf{x}, \mathbf{x} \, ; \mathbf{y}, \mathbf{X}, \mathbf{Z} ) \, d \mathbf{x}
    ,
    \label{eq:switching_criterion}
\end{equation}
where $c_{\text{ratio}}>0$ is a user-defined constant.
This criterion implicitly determines the number $M$ of inducing points required to adequately represent the posterior given the training dataset $\mathcal{D}$.
The choice $c_{\text{ratio}} =1$ corresponds to balancing the two IMSE variance contributions.
This parameter may also be adjusted to reflect different computational costs associated with training-data acquisition and inducing-input infilling. 
In the numerical experiments, $c_{\text{ratio}} =1$ yielded consistently satisfactory results, and no further tuning was performed.
In practice, the integrals in \cref{eq:switching_criterion} are estimated by Monte Carlo sampling.


\subsubsection{Pseudocode} \label{subsubsec:pseudo_code}
The sequential algorithm is summarized in \cref{alg:sequential_quantile_regression}.
Each update of the posterior quantile model involves solving the bilevel optimization problem \eqref{eq:bilevel_optimization_problem}.
The rejection sampling step is embarrassingly parallel, since candidate locations are drawn independently.

\begin{algorithm}
    \caption{Sequential sparse Gaussian process quantile regression}
    \label{alg:sequential_quantile_regression}
    \begin{algorithmic} 
        \STATE{Define initial training values $\mathbf{y}^{(0)}$, training inputs $\mathbf{X}^{(0)}$, and inducing inputs $\mathbf{Z}^{(0)}$.}
        \STATE{Train the posterior quantile model $q_\tau \mid \mathbf{y}^{(0)}, \mathbf{X}^{(0)}, \mathbf{Z}^{(0)}$.}
        \STATE{Set the iteration counters: $S \gets 0$, $T \gets 0$.}
        \WHILE{computational budget not exhausted}
            \IF{switching criterion not satisfied}
                \STATE{Set the batch size $B$.}
                \STATE{Sample $B$ training input locations $\mathbf{X}_{\text{new}} \in \mathcal{X}^B$ adaptively.}
                \STATE{Acquire observations $\mathbf{y}_{\text{new}} \in \mathbb{R}^B$ of the stochastic process $y ( \cdot )$ at $\mathbf{X}_{\text{new}}$.}
                \STATE{Augment the training dataset: $\mathbf{y}^{(S+1)} \gets [ \mathbf{y}^{(S)} ; \mathbf{y}_{\text{new}} ]$, $\mathbf{X}^{(S+1)} \gets [ \mathbf{X}^{(S)}, \mathbf{X}_{\text{new}} ]$.}
                \STATE{Update the posterior quantile model to $q_\tau \mid \mathbf{y}^{(S+1)}, \mathbf{X}^{(S+1)}, \mathbf{Z}^{(T)}$.}                
                \STATE{Increment the iteration counter: $S \gets S + 1$.}
            \ELSE
                \STATE{Maximize the infilling criterion for $\mathbf{z}_{\text{new}} \in \mathcal{X}$.}
                \STATE{Augment the inducing-input set: $\mathbf{Z}^{(T+1)} \gets [ \mathbf{Z}^{(T)}, \mathbf{z}_{\text{new}} ]$.}                
                \STATE{Update the posterior quantile model to $q_\tau \mid \mathbf{y}^{(S)}, \mathbf{X}^{(S)}, \mathbf{Z}^{(T+1)}$.}                
                \STATE{Increment the iteration counter: $T \gets T + 1$.}
            \ENDIF
        \ENDWHILE
        
        \RETURN the posterior quantile model $q_\tau \mid \mathbf{y}^{(S)}, \mathbf{X}^{(S)}, \mathbf{Z}^{(T)}$
    \end{algorithmic}
\end{algorithm}


\subsection{Computational considerations} \label{subsec:computational_considerations}
For a fixed inducing-input set $\mathcal{Z}$, training the posterior quantile model amounts to solving the bilevel problem \eqref{eq:bilevel_optimization_problem}.
The inner problem optimizes the $M$ inducing variables $\mathbf{u}$ for fixed hyperparameters, and the posterior covariance is recovered from the Hessian of $\widetilde{\mathcal{J}}_N$ at the MAP estimate rather than optimized.
The number of optimization variables associated with the inducing representation is thus $\mathcal{O} ( M )$, against $\mathcal{O} ( M^2 )$ for sparse variational formulations with a full covariance matrix.
The closed-form gradient and Hessian of $\widetilde{\mathcal{J}}_N$ further enable Newton's method.
For given hyperparameters, evaluating the conditional-prior mean and variance at the $N$ training inputs costs $\mathcal{O} ( N M ( M + D ) + M^3 )$.
Each Newton iteration then requires $\mathcal{O} ( N M )$ operations to evaluate the objective and its gradient, $\mathcal{O} ( N M^2 )$ to assemble the Hessian, and $\mathcal{O} ( M^3 )$ to factorize it, hence $\mathcal{O} ( N M^2 + M^3 )$ overall.
The gradient of $\hat{\mathcal{F}}_N$ with respect to the $D + 2$ hyperparameters can be obtained at the same order of cost as the conditional-prior quantities by adjoint differentiation.
The total training cost then scales with the numbers of inner and outer iterations.

Existing sparse GP approaches to quantile regression prescribe the number $M$ of inducing points \emph{a priori}.
Their locations are either fixed by predefined allocation strategies, such as random sampling, quasi-Monte Carlo (QMC) sequences, or $k$-means clustering, or optimized jointly with the variational parameters, which adds $M D$ optimization variables.
In contrast, the infilling criterion of \cref{subsec:inducing_input_infilling}, combined with the switching criterion \eqref{eq:switching_criterion}, determines both the number and the locations of the inducing points.
Inducing points are added only when the conditional-prior variance dominates the predictive uncertainty.
This keeps the representation parsimonious, so that training and prediction costs grow with $M$ only as required by the data, and avoids prescribing a large $M$ beforehand to guarantee accuracy.

This adaptivity comes at the cost of optimizing the infilling criterion.
Let $\mathbf{X}_{\mathrm{int}}$ denote the $N_{\mathrm{int}}$ points used to estimate the integral in \eqref{eq:integrated_variance_reduction} by Monte Carlo sampling.
The kernel matrix $K ( \mathbf{Z}, \mathbf{X}_{\mathrm{int}} )$ does not depend on the candidate $\tilde{\mathbf{z}}$ and can be computed once per infilling step, at a cost of $\mathcal{O} ( M N_{\mathrm{int}} D )$.
Using the factorization of $\Kzz$ available from training, each evaluation of the criterion and of its gradient then costs $\mathcal{O} ( ( N_{\mathrm{int}} + M ) ( M + D ) )$, independently of $N$ and without evaluating the stochastic process.
This cost is small compared with a Newton iteration when $( N_{\mathrm{int}} + M ) ( M + D ) \ll M^2 ( N + M )$, and negligible compared with the cost of the expensive simulators targeted by the present work.


\section{Numerical experiments} \label{sec:numerical_experiments}


\subsection{Numerical setup} \label{subsec:numerical_setup}
Throughout the experiments, the training input locations $\mathbf{X}$ are initialized uniformly over the input space $\mathcal{X}$.
The inducing inputs $\mathbf{Z}$ are initialized with a Halton sequence, a QMC sampling method.
Following Le Riche and Picheny~\cite{leriche2021}, we rescale the inputs to $[ 0, 1 ]^D$, standardize the outputs, and bound the lengthscales $\ell_d$ within $[ \sqrt{D}/100, \sqrt{D} ]$.
The signal variance $\sigma_{\text{s}}^2$ and the asymmetric Laplace scale parameter $\alpha$ are bounded within $[ 10^{-3}, 10^3 ]$ and $[ 10^{-3}, 10 ]$, respectively, and a jitter of $10^{-6}$ is added to the diagonal of $\Kzz$.
The outer optimization problem \eqref{eq:upper_level_optimization_problem} is solved by the L-BFGS algorithm~\cite{nocedal1980,liu1989} in logarithmic coordinates with five starts.
Four starts are drawn uniformly within the bounds, while the fifth is set to unit parameter values at the initial fit and to the preceding fit thereafter.
The strictly concave inner problem \eqref{eq:lower_level_optimization_problem} is solved in whitened coordinates by Newton's method, using the closed-form Hessian and a single start initialized from the preceding MAP estimate after the initial fit.
Relative tolerances on the parameters are $10^{-6}$ and $10^{-9}$ for the outer and inner problems, respectively.
The same settings are used in all experiments.


\subsection{Reporting conventions} \label{subsec:reporting_conventions}
From \cref{subsec:assessment_inducing_input_infilling} onward, the experiments are repeated over $64$ independent training datasets.
Solid, dashed, and dotted lines indicate averages over the independent runs, while shaded regions represent the corresponding $95\%$ confidence intervals.
The reported results are normalized by
\begin{equation}
    \epsilon_{{\tau, \text{true}}}
    =
    \int_{\mathcal{X}} \big( q_{\tau, \text{true}} ( \mathbf{x} ) - \bar{q}_{\tau, \text{true}} \big)^2 \, d \mathbf{x}
    ,
\end{equation}
where $\bar{q}_{\tau, \text{true}}$ denotes the mean of $q_{\tau, \text{true}}$ over $\mathcal{X}$.
The true conditional $\tau$-quantile function $q_{\tau, \text{true}}$ is derived analytically for the synthetic benchmarks of \cref{subsec:proof_of_concept}, and estimated empirically by Monte Carlo sampling for the application of \cref{subsec:application}.
For the synthetic benchmarks, the reported IMSE follows \cref{eq:imse_developed_form}.
For the application, only the squared error of the predictive mean is retained, since the two compared methods rest on different noise models and their predictive variances are not directly comparable.


\subsection{Proof of concept} \label{subsec:proof_of_concept}


\subsubsection{Benchmark functions} \label{subsec:benchmark_functions}
The performance and robustness of the proposed quantile regression method are assessed on two benchmark functions drawn from the literature.
These functions present a range of structural complexities, including nonlinearity, multimodality, and heteroskedasticity, which are characteristic of real-world modeling challenges.
\begin{itemize}    
    \item \textit{Sabater}.
    The first benchmark is the function introduced by Sabater et al.~\cite{sabater2021} to evaluate their Bayesian quantile optimization framework.
    The stochastic process is defined as
    \begin{equation}
        f ( \mathbf{x}, \omega )
        =
        \sum_{d=1}^{D} f ( \mathrm{x}_d, \omega )
        ,        
        \label{eq:sabater_function}
    \end{equation}
    where $\mathbf{x} \in [a, b]^D$ with $a = 2$ and $b = 8$, $\omega \in \varOmega$ denotes an outcome, and
    \begin{equation}
        f ( \mathrm{x}, \omega )
        =
        3 - 4 \exp \! \left( -4 ( \mathrm{x} - 4 )^2 \right) - 5.2 \exp \! \left( -4 ( \mathrm{x} - 6 )^2 \right) + \frac{ \mathrm{x} - a }{ b - a } \, \xi_1 ( \omega ) + \frac{ b - \mathrm{x} }{ b - a } \, \xi_2 ( \omega )
        .
    \end{equation}    
    The random variables $\xi_1 : \varOmega \to \mathbb{R}$ and $\xi_2 : \varOmega \to \mathbb{R}$ are independent and distributed as
    \begin{equation}
        \xi_1 \sim \mathcal{U} \big( 0, 10 \big)
        ,
        \quad
        \xi_2 \sim \mathcal{N} \big( 1.01, 0.71^2 \big)
        .
        \label{eq:sabater_uncertainty}
    \end{equation}
    The stochastic process exhibits mild heteroskedasticity and asymmetric noise, resulting from the distinct distributions of the two independent random variables.
    Consistent with the original setup, we evaluate performance at the $0.8$ quantile level.
    The $0.8$-quantile function has $2^D$ local minima.
    \item \textit{Michalewicz}.
    The second benchmark is a one-dimensional function used by Torossian et al.~\cite{torossian2020} to evaluate quantile regression methods.
    The stochastic process, based on the Michalewicz function~\cite{dixon1978}, is defined as
    \begin{equation}
        f ( \mathrm{x}, \omega )
        =
        -2 \sin ( \mathrm{x} ) \sin^{30} \! \left( \frac{\mathrm{x}^2}{\pi} \right) 
        - \frac{0.1 \cos^3 \! \left( \frac{\pi \mathrm{x}}{10} \right) }{\left\lvert 2 - \sin ( \mathrm{x} ) \sin^{30} \! \left( \frac{\mathrm{x}^2}{\pi} \right) \right\rvert } \, \xi^2 ( \omega )
        ,
        \label{eq:michalewicz_function}
    \end{equation}
    where $\mathrm{x} \in [0, 4]$, $\omega \in \varOmega$, and $\xi : \varOmega \to \mathbb{R}$ is given by
    \begin{equation}
        \xi ( \omega ) = 3 \eta ( \omega ) \, \mathds{1}_{ \{ \eta ( \omega ) < 0 \} } + 6 \eta ( \omega ) \, \mathds{1}_{ \{ \eta ( \omega ) \geq 0 \} }
        ,
        \quad \text{with} \quad
        \eta \sim \mathcal{N} \big( 0, 1 \big)
        .
        \label{eq:michalewicz_uncertainty}
    \end{equation}
    The stochastic process is characterized by pronounced heteroskedasticity and an asymmetric noise structure.
    Performance is evaluated at the $0.9$ quantile level, one of the two extreme levels considered in the original study.
    The $0.9$-quantile function exhibits sharply localized shape variations and is nearly flat elsewhere.
\end{itemize}


\subsubsection{Empirical validity of the Laplace approximation} \label{subsec:empirical_validity_laplace_approximation}
This section assesses the Laplace approximation to the surrogate posterior $\widetilde{p}\propto\exp(\widetilde{\mathcal{J}}_N)$ introduced in \cref{subsec:approximate_inference}.
We compare this density with its Gaussian approximation \eqref{eq:inducing_variables_posterior_gaussian_approximation}, isolating the error of the Laplace approximation.
The assessment is carried out on the one-dimensional variant of the \textit{Sabater} function with $M = 10$ inducing inputs.
Three training datasets of sizes $N = 100$, $1{,}000$, and $10{,}000$ observations are considered, representing small, intermediate, and large training-dataset regimes for this benchmark.
For each dataset, the GP hyperparameters $\boldsymbol{\psi}$ and the asymmetric Laplace scale parameter $\alpha$ are fixed at the solution of the bilevel problem \eqref{eq:bilevel_optimization_problem}.

\begin{figure}
    \centering
    \includegraphics[width=0.98\textwidth]{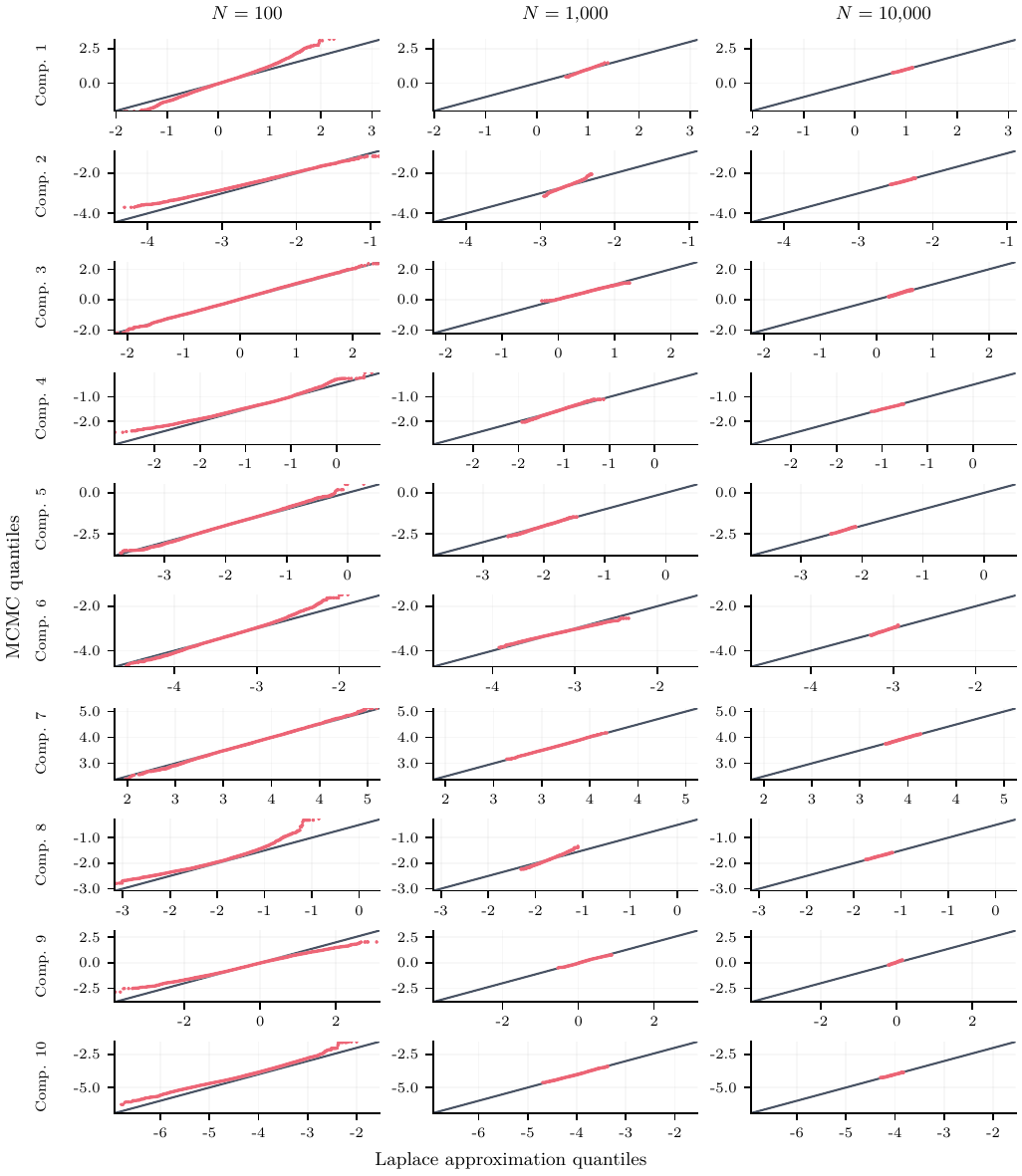}
    \caption{
        Q--Q plots comparing samples from the surrogate posterior $\widetilde{p} ( \mathbf{u} \mid \mathbf{Z}, \mathbf{y}, \mathbf{X} )$, obtained by Markov chain Monte Carlo (MCMC), with samples from its Laplace approximation, for each component $m = 1, \ldots, 10$ of the inducing variables.
        Results are shown for the \textit{Sabater 1D} benchmark function, using datasets with $N = 100$ (left), $1{,}000$ (middle), and $10{,}000$ observations (right).        
    }
    \label{fig:validation_mcmc_laplace_sabater_1d_qqplot}
\end{figure}

Our analysis proceeds in two steps.
First, we examine how agreement between the surrogate posterior and its Laplace approximation evolves with dataset size.
We compare the order statistics of $10^5$ MCMC samples from the surrogate posterior $\widetilde{p} ( \mathbf{u} \mid \mathbf{Z}, \mathbf{y}, \mathbf{X} )$ against the same number drawn from its Laplace approximation $\hat{p} ( \mathbf{u} \mid \mathbf{Z}, \mathbf{y}, \mathbf{X} )$, for each dataset size.
The resulting Q--Q plots, one per component $m = 1, \ldots, M$ of the multivariate distribution, are reported in \cref{fig:validation_mcmc_laplace_sabater_1d_qqplot}.

As \cref{fig:validation_mcmc_laplace_sabater_1d_qqplot} shows, agreement between the two sets of order statistics improves consistently across all components as the dataset size $N$ increases.
This indicates that the surrogate posterior becomes increasingly well approximated by a Gaussian distribution.
For the largest dataset ($N = 10{,}000$), the two distributions appear nearly indistinguishable, while for the intermediate dataset ($N = 1{,}000$), the approximation is already reasonably accurate.
Noticeable discrepancies remain only for the smallest dataset ($N = 100$), which suggests that the Laplace approximation is less reliable in low-data regimes.

While the Q--Q plots capture marginal agreement in the inducing variables, they do not reflect potential discrepancies in the joint structure.
Thus, the second analysis compares the posterior-induced variance $\Sigma_u^2 ( \mathbf{x}, \mathbf{x} \, ; \mathbf{y}, \mathbf{X}, \mathbf{Z} )$ across the input space $\mathcal{X}$, using the covariance matrix $\hat{\mathbf{C}}$ obtained either from the Laplace approximation or estimated empirically from MCMC samples.
Predictions are evaluated on a uniform grid of $10{,}000$ points.

\begin{figure}
    \centering
    \includegraphics[width=\textwidth]{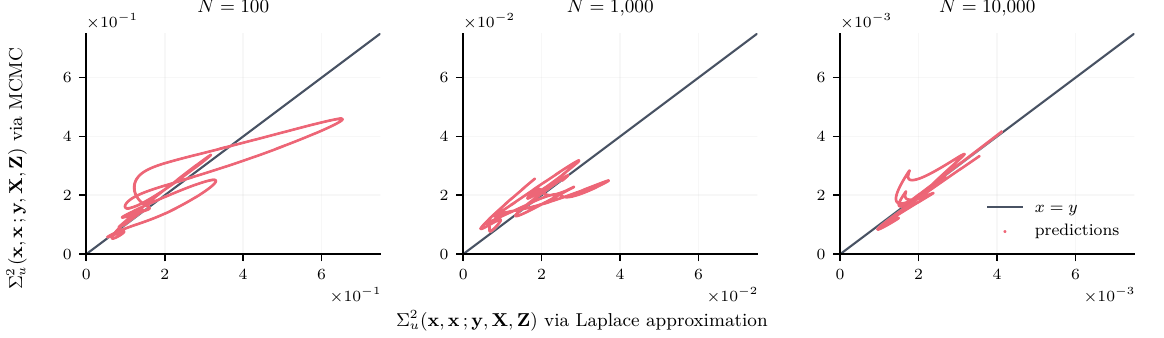}
    \caption{
        Comparison of posterior-induced variance predictions obtained with the covariance of the surrogate posterior estimated by Markov chain Monte Carlo (MCMC) and by the Laplace approximation.
        Results are shown for the \textit{Sabater 1D} benchmark function, using datasets with $N = 100$ (left), $1{,}000$ (middle), and $10{,}000$ observations (right).
    }
    \label{fig:validation_mcmc_laplace_sabater_1d_posterior_induced_variance_predictions}
\end{figure}

\Cref{fig:validation_mcmc_laplace_sabater_1d_posterior_induced_variance_predictions} shows improving agreement with MCMC as $N$ grows, supporting the accuracy of the Laplace approximation to the surrogate posterior over the tested dataset sizes.


\subsubsection{Inducing-input infilling} \label{subsec:assessment_inducing_input_infilling}
With a sufficiently large and well-chosen inducing-input set $\mathcal{Z}$, quantile predictions should approach the exact solution, provided that enough training data are available.
This section examines how our integrated variance reduction (IVR) infilling strategy compares with predefined inducing-input allocation in this regime.

\begin{figure}
    \centering
    \begin{subfigure}{0.49\textwidth}
        \centering
        \includegraphics[width=\textwidth]{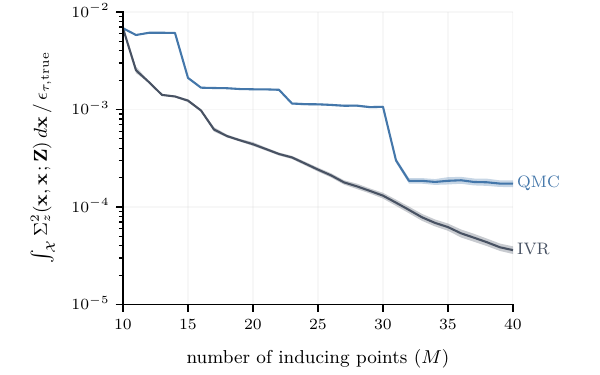}
        \caption{Integrated conditional-prior variance.}
        \label{subfig:inducing_input_infilling_sabater_1d_integrated_conditional_prior_variance}
    \end{subfigure}
    \hfill
    \begin{subfigure}{0.49\textwidth}
        \centering
        \includegraphics[width=\textwidth]{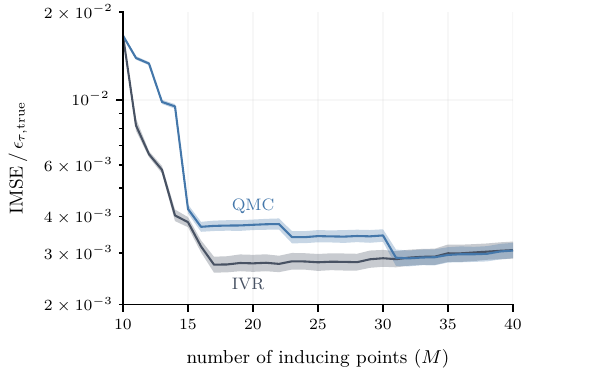}
        \caption{Integrated mean squared error.}
        \label{subfig:inducing_input_infilling_sabater_1d_imse}
    \end{subfigure}    
    \caption{
        Evolution of the normalized (a) integrated conditional-prior variance and (b) integrated mean squared error (IMSE) as functions of the number $M$ of inducing points, for the adaptive inducing-input infilling strategy (IVR, black) and quasi-Monte Carlo sampling (QMC, blue).                
        Results are shown for the \textit{Sabater 1D} benchmark function.
    }
    \label{fig:inducing_input_infilling_sabater_1d}
\end{figure}

QMC sampling via Halton sequences serves as the reference predefined inducing-input allocation strategy.
Both infilling strategies are first evaluated on the one-dimensional version of the \textit{Sabater} function with $N = 10{,}000$ observations.
The inducing-input set is initialized with $M = 10$ to give the initial quantile model sufficient structural complexity to guide the adaptive infilling.
\Cref{subfig:inducing_input_infilling_sabater_1d_integrated_conditional_prior_variance} tracks the evolution of the conditional-prior variance $\Sigma_z^2 ( \mathbf{x}, \mathbf{x} \, ; \mathbf{Z} )$, integrated over the input space $\mathcal{X}$, as each infilling strategy progresses.
\Cref{subfig:inducing_input_infilling_sabater_1d_imse} illustrates how this is reflected in predictive accuracy, as measured by the IMSE.

The \textit{Sabater 1D} function is one-dimensional, smooth, and moderately multimodal.
As a result, only a few inducing points are needed to capture its structure.
As shown in \cref{subfig:inducing_input_infilling_sabater_1d_integrated_conditional_prior_variance}, the integrated conditional-prior variance decays exponentially under adaptive infilling, outpacing that achieved by the Halton sequence.
This translates into a corresponding IMSE reduction, as seen in \cref{subfig:inducing_input_infilling_sabater_1d_imse}, until the error plateaus near $M = 17$ inducing points.
On average over the $64$ independent runs, the adaptive infilling strategy outperforms the Halton sequence in terms of IMSE up to approximately $M = 31$, beyond which both strategies achieve comparable predictive accuracy.
The integrated conditional-prior variance continues to decay exponentially even after the IMSE has leveled off.
This shows that the adaptive infilling strategy remains robust once further reductions in conditional-prior variance no longer improve predictions.
Such a decoupling suggests that the remaining predictive error is dominated by sources not addressed by the infilling criterion.

\begin{figure}
    \centering
    \includegraphics[width=0.49\textwidth]{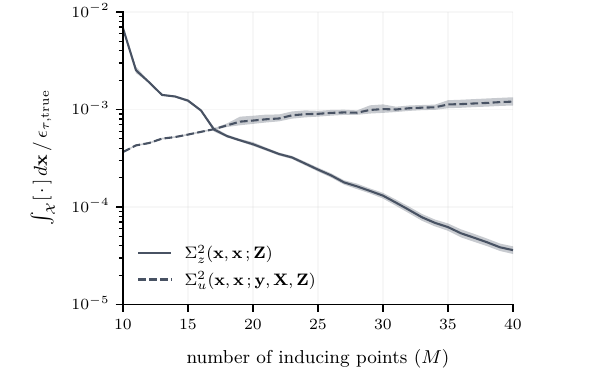}
    \caption{
        Evolution of the normalized integrated conditional-prior variance (solid line) and integrated posterior-induced variance (dashed line) as functions of the number $M$ of inducing points.        
        Results are shown for the \textit{Sabater 1D} benchmark function.
    }
    \label{fig:inducing_input_infilling_sabater_1d_integrated_variance_comparison}
\end{figure}

\Cref{fig:inducing_input_infilling_sabater_1d_integrated_variance_comparison} examines this transition by illustrating how the two IMSE variance components evolve under adaptive infilling.
These components are the conditional-prior variance $\Sigma_z^2 ( \mathbf{x}, \mathbf{x} \, ; \mathbf{Z} )$ and the posterior-induced variance $\Sigma_u^2 ( \mathbf{x}, \mathbf{x} \, ; \mathbf{y}, \mathbf{X}, \mathbf{Z} )$, both integrated over the input space $\mathcal{X}$.
Both curves intersect near $M = 17$ inducing points, coinciding with the onset of the IMSE plateau in \cref{subfig:inducing_input_infilling_sabater_1d_imse}.
Beyond this threshold, the variance induced by posterior uncertainty in the inducing variables emerges as the dominant error contribution.
Further infilling can therefore no longer reduce the IMSE substantially, and it becomes more effective to stop the inducing-input infilling and trigger data acquisition.
The integrated variance terms analyzed here correspond precisely to the left- and right-hand sides of the switching criterion \eqref{eq:switching_criterion}.
As a result, the proposed switching criterion would halt the inducing-input infilling within the regime of significant IMSE reduction and trigger data acquisition accordingly.
It has proved effective in all our experiments.

\begin{figure}
    \centering
    \begin{subfigure}{0.49\textwidth}
        \centering
        \includegraphics[width=\textwidth]{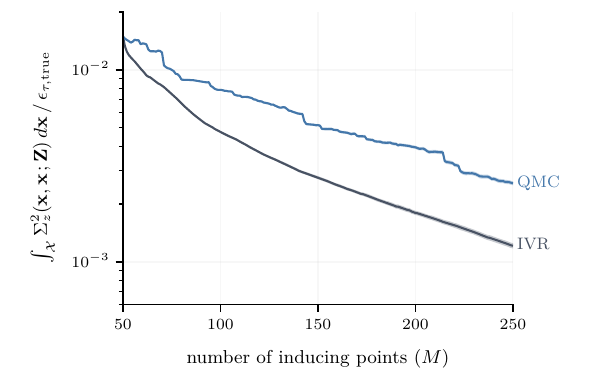}
        \caption{Integrated conditional-prior variance.}
        \label{subfig:inducing_input_infilling_sabater_2d_integrated_conditional_prior_variance}
    \end{subfigure}
    \hfill
    \begin{subfigure}{0.49\textwidth}
        \centering
        \includegraphics[width=\textwidth]{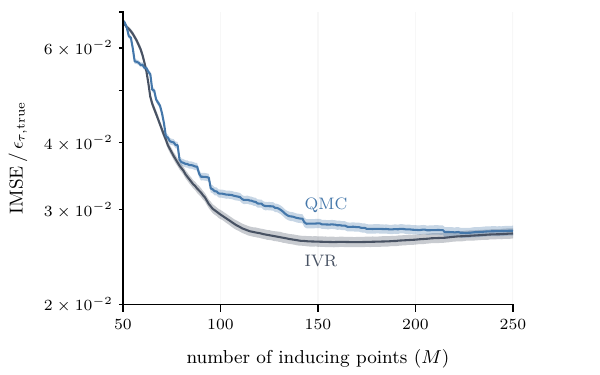}
        \caption{Integrated mean squared error.}
        \label{subfig:inducing_input_infilling_sabater_2d_imse}
    \end{subfigure}    
    \caption{
        Evolution of the normalized (a) integrated conditional-prior variance and (b) integrated mean squared error (IMSE) as functions of the number $M$ of inducing points, for the adaptive inducing-input infilling strategy (IVR, black) and quasi-Monte Carlo sampling (QMC, blue).                
        Results are shown for the \textit{Sabater 2D} benchmark function.
    }
    \label{fig:inducing_input_infilling_sabater_2d}
\end{figure}

We next examine inducing-input infilling in higher dimension by reproducing the above experiments on the two-dimensional variant of the \textit{Sabater} function.
The quantile model is initialized with $M=50$ inducing points and $N=10{,}000$ observations.
The integrated conditional-prior variance and the IMSE are reported in \cref{fig:inducing_input_infilling_sabater_2d}.

In line with the earlier findings, IVR consistently outperforms the predefined inducing-input allocation strategy in terms of integrated conditional-prior variance.
The exponential decay observed in the one-dimensional case also holds in higher dimension.
In this example, the IMSE achieved through IVR does not plateau but instead reaches a minimum before rising again.
The minimum occurs near $M=140$ inducing points.
Compared with the one-dimensional case, this indicates that more inducing points are needed to capture the increased structural complexity of higher-dimensional settings.
The IMSE deterioration occurs because the available training data become insufficient to reliably estimate the inducing variables, causing the posterior-induced variance contribution to increase.
As established previously, the switching criterion \eqref{eq:switching_criterion} would guard against such a deterioration.


\subsubsection{Predictive accuracy with different training dataset sizes} \label{subsec:predictive_accuracy_with_different_dataset_sizes}
The previous experiments show that our adaptive inducing-input infilling strategy outperforms predefined allocation when sufficient training data are available.
It remains well-behaved throughout the infilling, even when the targeted variance contribution in the IMSE becomes dominated by other contributions.
This section investigates how the global predictive accuracy of the quantile model varies with the number $N$ of observations in the training dataset $\mathcal{D}$.
We study how the optimal number of inducing points evolves with increasing data availability, and how the model behaves under limited data regimes.

\begin{figure}
    \centering
    \includegraphics[width=0.49\textwidth]{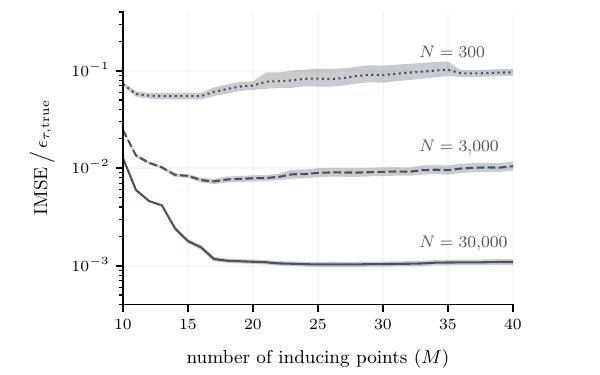}    
    \caption{
        Evolution of the normalized integrated mean squared error (IMSE) as a function of the number $M$ of inducing points, for training dataset sizes $N = 300$ (dotted line), $3{,}000$ (dashed line), and $30{,}000$ (solid line).                
        Results are shown for the \textit{Sabater 1D} benchmark function.
    }
    \label{fig:different_obs_sabater_1d}
\end{figure}

Three training datasets of sizes $N = 300$, $3{,}000$, and $30{,}000$ are considered for the one-dimensional \textit{Sabater} benchmark.
We compare the evolution of the IMSE throughout the adaptive inducing-input infilling for each dataset size, starting from $M = 10$ inducing points.
Results are reported in \cref{fig:different_obs_sabater_1d}.

As shown in \cref{fig:different_obs_sabater_1d}, the number of inducing points required to reach the minimum IMSE increases with the number of observations.
This indicates that larger datasets can support more complex inducing representations.
Moreover, the minimum IMSE decreases as $N$ grows, confirming that the best achievable predictive accuracy improves with more data.
In the asymptotic regime $N \to \infty$, the complexity of the true $\tau$-quantile function can be fully captured, and predictions should converge to the exact values.
These observations suggest that the proposed strategy is compatible with asymptotic convergence toward the true quantile function as both the training-set size and the inducing-point complexity increase.
With limited training data ($N = 300$ and $3{,}000$), the IMSE reaches a minimum before rising again, for the same reason as in the two-dimensional example of \cref{subsec:assessment_inducing_input_infilling}.

We then assess the adaptive control of $M$ provided by the switching criterion \eqref{eq:switching_criterion}.
For each run, we record the number of inducing points at which the criterion stops the infilling.
\Cref{tab:imse_optimal_inducing_count} compares its average over the runs with the empirical IMSE minimizer, defined as the single $M$ that minimizes the run-averaged normalized IMSE.
The resulting loss in accuracy is quantified by the relative excess of the normalized IMSE at the selected number of inducing points over that at the empirical minimizer.

\begin{table}
    \centering
    \caption{
        Number of inducing points $M$ selected by the switching criterion \eqref{eq:switching_criterion} compared with the empirical integrated mean squared error (IMSE) minimizer, for training dataset sizes $N = 300$, $3{,}000$, and $30{,}000$.
        Results are shown for the \textit{Sabater 1D} benchmark function.
    }
    \label{tab:imse_optimal_inducing_count}
    \begin{adjustbox}{max width=\textwidth}
        \begin{tabular}{@{}rccccc@{}}
            \toprule
            & \multicolumn{2}{c}{Empirical minimizer} & \multicolumn{3}{c}{Switching criterion} \\
            \cmidrule(lr){2-3} \cmidrule(l){4-6}
            $N$ & $M$ & $\text{IMSE} / \epsilon_{{\tau, \text{true}}}$ & $M$ & $\text{IMSE} / \epsilon_{{\tau, \text{true}}}$ & excess \\
            \midrule
            $300$ & $14$ & $5.48 \times 10^{-2}$ & $11.1$ & $5.81 \times 10^{-2}$ & $6.2\%$ \\
            $3{,}000$ & $17$ & $7.32 \times 10^{-3}$ & $15.6$ & $7.82 \times 10^{-3}$ & $6.8\%$ \\
            $30{,}000$ & $25$ & $1.03 \times 10^{-3}$ & $21.0$ & $1.09 \times 10^{-3}$ & $5.4\%$ \\
            \bottomrule
        \end{tabular}
    \end{adjustbox}
\end{table}

The number of inducing points selected by the switching criterion increases with $N$, as does the empirical minimizer, while remaining below it on average.
The resulting excess in normalized IMSE is at most $6.8\%$, which is small compared with the reduction in normalized IMSE by a factor of about $50$ between $N = 300$ and $N = 30{,}000$.
Unlike the empirical minimizer, which requires the true $\tau$-quantile, the criterion relies only on variance terms computed by the model itself.
It can thus be considered an effective heuristic to adaptively control the number of inducing points.


\subsubsection{Sequential enrichment} \label{subsec:assessment_sequential_enrichment}
\Cref{subsec:assessment_inducing_input_infilling} shows that infilling the inducing-input set by maximizing the integrated reduction in conditional-prior variance outperforms predefined allocation, provided sufficient training data.
\Cref{subsec:predictive_accuracy_with_different_dataset_sizes} shows that global predictive accuracy improves with training dataset size and that the optimal number of inducing points grows accordingly.
The variance-based switching criterion \eqref{eq:switching_criterion} is designed to identify the dominant variance contribution in the IMSE, and to trigger enrichment of either the inducing-input set $\mathcal{Z}$ or the training dataset $\mathcal{D}$ accordingly.
This section compares the convergence of the sequential algorithm, which comprises inducing-input infilling and training-data acquisition, as a function of training dataset size, for two acquisition strategies: uniform sampling and the rejection sampling scheme introduced in \cref{subsec:training_data_acquisition}.

\begin{figure}
    \centering
    \begin{subfigure}{0.49\textwidth}
        \centering
        \includegraphics[width=\textwidth]{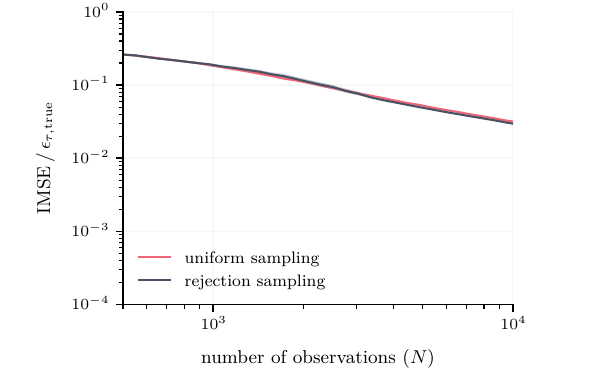}
        \caption{\textit{Sabater 2D}}
        \label{fig:different_query_sabater_2d}
    \end{subfigure}
    \hfill
    \begin{subfigure}{0.49\textwidth}
        \centering
        \includegraphics[width=\textwidth]{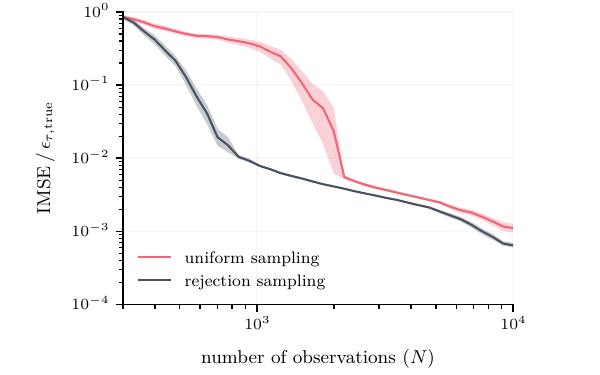}
        \caption{\textit{Michalewicz}}
        \label{fig:different_query_michalewicz}
    \end{subfigure}
    \caption{
        Evolution of the normalized integrated mean squared error (IMSE) as a function of the number $N$ of observations, for the sequential algorithm with data acquired through rejection sampling (black) and uniform sampling (red).
        Results are shown for (a) the \textit{Sabater 2D} and (b) the \textit{Michalewicz} benchmark functions.
    }
    \label{fig:different_query}
\end{figure}

The comparison is carried out on both the \textit{Sabater 2D} and \textit{Michalewicz} functions.
For the \textit{Sabater 2D} benchmark, the initial numbers of observations and inducing points are $N = 500$ and $M = 50$, respectively.
The \textit{Michalewicz} benchmark is initialized with $N = 300$ observations and $M = 10$ inducing points.
In both cases, subsequent inducing inputs are added through IVR.
Training data are acquired in batches equal to $10\%$ of the current dataset.
Results are reported in \cref{fig:different_query}.

For the \textit{Sabater 2D} benchmark (\cref{fig:different_query_sabater_2d}), rejection sampling performs on par with uniform sampling.
This suggests that, for this function, the benefits of adaptivity are concentrated in the inducing-input allocation rather than in the data-acquisition strategy.
In contrast, for the \textit{Michalewicz} benchmark (\cref{fig:different_query_michalewicz}), rejection sampling yields substantially lower IMSE values than uniform sampling up to approximately $2{,}000$ observations, showing that adaptive data acquisition can provide significant gains over predefined strategies, even for one-dimensional functions.
The \textit{Michalewicz} benchmark exhibits localized shape variations that rejection sampling helps discover early.
This early discovery then guides the inference toward predictions that significantly outperform those obtained with uniform sampling, until the latter has acquired enough data for the model to recover the predictive accuracy of the rejection-based model.
We therefore conclude that adaptive data acquisition is particularly beneficial in data-limited settings for functions with localized shape variations.
Overall, in both cases, the IMSE decreases consistently as the algorithm progresses, in line with the asymptotic behavior discussed in \cref{subsec:predictive_accuracy_with_different_dataset_sizes}.

\begin{figure}
    \centering
    \begin{subfigure}{0.49\textwidth}
        \centering
        \includegraphics[width=\textwidth]{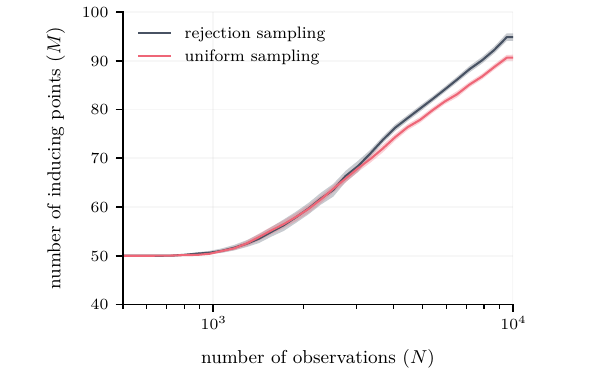}
        \caption{\textit{Sabater 2D}}
        \label{fig:different_query_sabater_2d_inducing}
    \end{subfigure}
    \hfill
    \begin{subfigure}{0.49\textwidth}
        \centering
        \includegraphics[width=\textwidth]{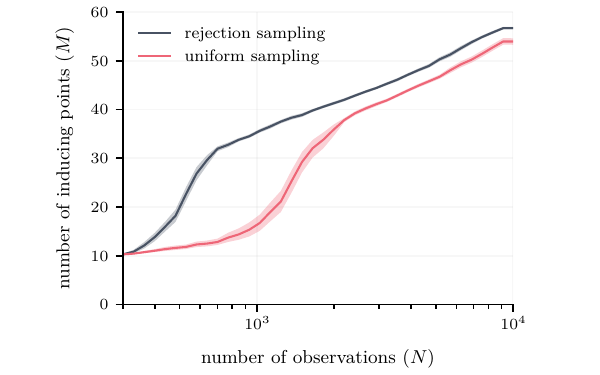}
        \caption{\textit{Michalewicz}}
        \label{fig:different_query_michalewicz_inducing}
    \end{subfigure}
    \caption{
        Evolution of the number $M$ of inducing points as a function of the number $N$ of observations, for the sequential algorithm with data acquired through rejection sampling (black) and uniform sampling (red).
        Results are shown for (a) the \textit{Sabater 2D} and (b) the \textit{Michalewicz} benchmark functions.
    }
    \label{fig:different_query_inducing}
\end{figure}

\Cref{fig:different_query_sabater_2d_inducing,fig:different_query_michalewicz_inducing} report the number $M$ of inducing points selected by the switching criterion along the same runs.
On both benchmarks, $M$ grows with the number $N$ of observations, but at a much slower rate.
On the \textit{Sabater 2D} benchmark, $M$ remains at its initial value of $50$ until $N \approx 1{,}000$ and then rises to about $90$--$95$ at $N = 10{,}000$, for both uniform and rejection sampling.
The initial representation is thus sufficient for the first observations.
On the \textit{Michalewicz} benchmark, $M$ rises from $10$ to about $55$ while $N$ grows from $300$ to $10{,}000$.
Rejection sampling increases $M$ earlier than uniform sampling on this benchmark, in line with its earlier discovery of the localized variations of the function.
The switching criterion therefore adapts the complexity of the quantile model to the information carried by the data, without $M$ being prescribed beforehand.


\subsection{Application} \label{subsec:application}
We apply our approach to the construction of a seakeeping operating envelope for a container ship under wave-forecast uncertainty.
A standard seakeeping criterion for crew comfort and cargo safety is the vertical acceleration at the bow~\cite{nordforsk1987}.
The operating envelope identifies the speed--heading conditions in which this acceleration remains below a given limit with a prescribed probability.

The vessel under study is the KRISO Container Ship (KCS)~\cite{kim2001}, a standard container-ship benchmark.
The controllable input is $\mathbf{x} = ( U, \beta ) \in \mathcal{X} = [12, 24]$ kn $\times [0^\circ, 180^\circ]$, where $U$ is the vessel speed and $\beta$ the heading relative to the mean wave direction, with $\beta = 0^\circ$ corresponding to head seas.
We define the response as the peak of the absolute vertical acceleration $a_{z, \mathrm{bow}}$ at a bow station over a half-hour interval:
\begin{equation}
    f ( \mathbf{x}, \omega )
    =
    \max_{0 \leq t \leq T}
    \left\lvert a_{z, \mathrm{bow}} ( t \, ; \mathbf{x}, \omega ) \right\rvert
    ,
    \qquad
    T = 30~\mathrm{min}
    .
    \label{eq:kcs_peak_acceleration}
\end{equation}
This response is random through the sea state, represented by $\omega$ and detailed below.
We estimate its conditional $0.9$-quantile $q_{0.9} ( \mathbf{x} )$.
For a prescribed acceleration limit $a_{\mathrm{lim}}$, the operating envelope then consists of the speed--heading conditions $\mathbf{x}$ satisfying $q_{0.9} ( \mathbf{x} ) \leq a_{\mathrm{lim}}$.

The acceleration history is synthesized from $J = 2{,}048$ regular-wave components with frequencies $\varpi_j$, spaced by $\Delta \varpi_j$ and spanning periods from $7$ to $30$ s,
\begin{equation}
    a_{z, \mathrm{bow}} ( t \, ; \mathbf{x}, \omega )
    =
    \operatorname{Re}
    \sum_{j=1}^{J}
    H_a ( \varpi_j \, ; U, \beta_r ( \omega ) )
    \sqrt{ 2 S ( \varpi_j \, ; H_s ( \omega ), T_p ( \omega ) ) \Delta \varpi_j } \,
    e^{ \mathrm{i} \left( \vartheta_j ( \omega ) - \varpi_e ( \varpi_j \, ; U, \beta_r ( \omega ) ) \, t \right) }
    ,
    \label{eq:kcs_acceleration_synthesis}
\end{equation}
where the $\vartheta_j$ are random phases, and $S$ is the long-crested JONSWAP spectrum~\cite{hasselmann1973} with significant wave height $H_s$, peak period $T_p$, and peak-enhancement factor $3.3$.
Each component oscillates at the encounter frequency $\varpi_e ( \varpi \, ; U, \beta ) = \varpi + \varpi^2 U \cos \beta / g$, where $g$ is the gravitational acceleration.
The history is sampled every $0.25$ s, and the maximum in \eqref{eq:kcs_peak_acceleration} is taken over these samples.
This time step and the number $J$ of components are selected by a convergence study at fixed sea states and headings.

The outcome $\omega$ determines $2{,}051$ random variables, namely $2{,}048$ \iid phases $\vartheta_j \sim \mathcal{U} [ 0, 2 \pi )$ representing the irregular sea, and three forecast errors, in wave height, period, and direction.
The height and period errors are independent and log-normal around a forecast with $\hat{H}_s = 3$ m and $\hat{T}_p = 14$ s.
We set $\log ( H_s / \hat{H}_s ) \sim \mathcal{N} ( 0, 0.13^2 )$ and $\log ( T_p / \hat{T}_p ) \sim \mathcal{N} ( 0, 0.09^2 )$, truncated to $H_s \in [ 1.5, 5 ]$ m and $T_p \in [ 12, 20 ]$ s.
The direction error $\Delta \beta \sim \mathcal{N} ( 0, ( 10^\circ )^2 )$ gives the realized heading $\beta_r = \beta + \Delta \beta$, folded into $[ 0^\circ, 180^\circ ]$ by port--starboard symmetry.
The standard deviations of the three errors are calibrated against $24$-hour ECMWF wave forecasts and observations from six Iberian Atlantic buoys over $2024$--$2026$.
The heading error makes the noise strongly heteroskedastic.

\begin{figure}
    \centering
    \includegraphics[width=0.8\textwidth]{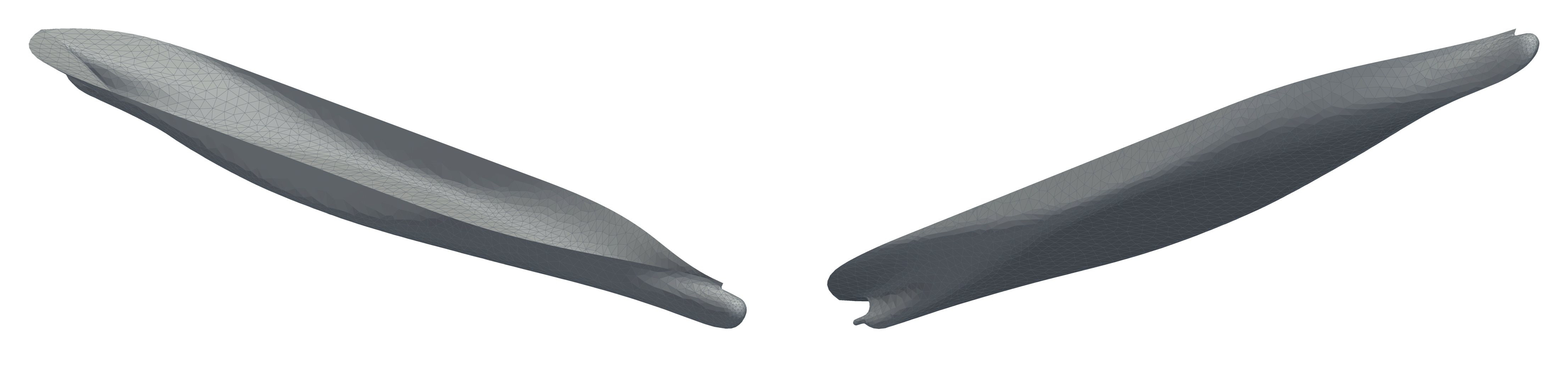}
    \caption{
        Upper (left) and lower (right) views of the wetted-hull mesh of the KRISO Container Ship (KCS) used in the boundary element computations ($4{,}732$ triangular panels, symmetric about the centerplane).
    }
    \label{fig:kcs_hull_mesh}
\end{figure}

The response-amplitude operator (RAO) $H_a$ maps each incident wave component to the bow vertical acceleration.
It is obtained from the heave and pitch responses of the hull, computed within linear potential-flow theory by the boundary element method (BEM) solver Capytaine with forward speed~\cite{babarit2015,ancellin2019,donatini2022}.
The wetted hull is discretized with a symmetric mesh of $4{,}732$ triangular panels, shown in \cref{fig:kcs_hull_mesh}.
Computing the RAO by BEM at one speed--heading condition, over all computed wave frequencies, takes about $40$ CPU core-minutes.
Since the response model is linear in the wave elevation, the RAO does not depend on the sea state, and each realization of $f$ reduces to the superposition \eqref{eq:kcs_acceleration_synthesis} once the RAO is known.
The heading error, however, implies that each realization, including replications, requires the RAO at its own realized heading.
A BEM solve per realization remains affordable in a practical application, but would be prohibitive for the present validation study, which involves a brute-force estimation of the reference quantile function and an assessment of the variability across independent runs.
The RAO is therefore replaced by a surrogate model fitted to BEM solutions at $192$ speed--heading conditions.
These solutions are compressed over the frequency into $13$ principal components, whose coefficients are interpolated over the speed--heading domain by independent GPs.
The same surrogate RAO is used for all runs and for the reference.

We compare our method with \textit{quantile kriging}~\cite{plumlee2014}, which interpolates by GP regression the empirical quantiles of $R$ realizations per site, with a correction based on the site means.
To obtain a sequential counterpart to our approach, new sites are selected one at a time by integrated variance reduction over $\mathcal{X}$.
We consider three replication levels, $R \in \{ 10, 25, 50 \}$.
Our approach uses acquisition batches of the same sizes, $B \in \{ 10, 25, 50 \}$, so that both methods consume the budget at the same rate.
It starts from $N = 100$ observations and $M = 10$ inducing points, and quantile kriging from $10$ sites placed at these inducing inputs, hence from $N = 10R$ realizations.
Both methods stop at $N = 2{,}000$.

We track the normalized IMSE as a function of the number $N$ of evaluations of $f$, averaged over $64$ independent runs.
The reference quantile function is estimated by Monte Carlo sampling on a $31 \times 31$ grid, with $8{,}192$ realizations per grid point and common random inputs across grid points.
Repeating this estimation with different random inputs yields a normalized IMSE between the two estimates about two orders of magnitude below the smallest value reached by either method.

\begin{figure}
    \centering
    \includegraphics[width=0.49\textwidth]{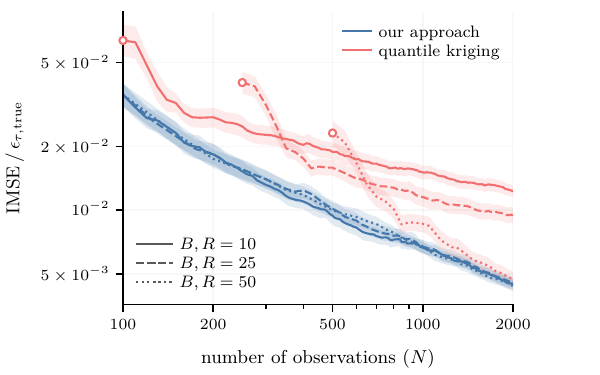}
    \caption{
        Evolution of the normalized integrated mean squared error (IMSE) as a function of the number $N$ of observations for the KCS application, for our sequential Bayesian approach (blue) with acquisition batches of size $B$ and quantile kriging (red) with $R$ replications per site.
        Solid, dashed, and dotted lines correspond to $B, R = 10$, $25$, and $50$, respectively.
        Open circles mark the first budget at which each design for quantile kriging is available ($N = 10R$).
    }
    \label{fig:application_nimse_comparison}
\end{figure}

\Cref{fig:application_nimse_comparison} compares the two methods.
At $N = 500$, the error of our approach is $1.6$ to $2.5$ times lower than that of any replication level.
At equal batch size and replication level ($B = R$), our approach yields a lower mean error at every common budget.
For $B = R = 10$ and $25$, the gap reaches a factor of $2.8$ and $2.1$ at $N = 2{,}000$, respectively.
For $B = R = 50$, the two methods perform comparably from about $N = 1{,}200$ onward.
Our approach is robust to the batch size, whereas quantile kriging is highly sensitive to the replication level.
At $N = 2{,}000$, $R = 10$ yields an error about $2.6$ times that for $R = 50$.
Our approach therefore outperforms quantile kriging at low and moderate budgets, and performs comparably to its best replication level at the largest budget without requiring this level to be chosen.


\section{Conclusion} \label{sec:conclusion}

This work introduced a sequential sparse Gaussian process framework for Bayesian quantile regression.
The proposed approach combines a sparse representation of the latent quantile function through inducing variables with a Laplace approximation to their posterior distribution.
To avoid the repeated sampling that would otherwise be required to evaluate the exact posterior objective, inference is recast as the optimization of a tractable surrogate log-posterior, yielding a sample-free Bayesian inference procedure together with closed-form expressions for the predictive mean, covariance, and their derivatives.
The resulting formulation leads to a reduced optimization space compared with sparse variational approaches and enables efficient deterministic, gradient-based optimization.

A central outcome of this work is the decomposition of the predictive uncertainty into two complementary contributions: a conditional-prior variance associated with the inducing representation and a posterior-induced variance associated with uncertainty in the inducing variables. 
This decomposition provides a principled framework for sequentially improving the predictive quantile model. 
The conditional-prior variance motivates an adaptive inducing-input infilling strategy that progressively enriches the sparse representation of the quantile function. 
The posterior-induced variance, in turn, drives an adaptive data-acquisition procedure that focuses on regions where uncertainty in the inducing variables has the largest impact on the quantile predictions.
These two enrichment mechanisms are unified through a variance-based switching criterion that balances model-complexity growth and data acquisition according to the dominant source of predictive uncertainty.

The numerical experiments on synthetic benchmarks demonstrate several important properties of the proposed methodology.
First, the Laplace approximation provides accurate estimates of the posterior-induced covariance when compared with reference Markov chain Monte Carlo computations.
Second, the variance-based inducing-input infilling outperforms predefined allocation strategies in terms of conditional-prior variance reduction and predictive accuracy.
Third, the evolution of the two variance contributions confirms the rationale of the proposed switching criterion, which identifies the regime where further enrichment of the inducing representation yields diminishing returns and additional data acquisition becomes preferable.
Finally, the complete sequential strategy improves predictive accuracy while adaptively controlling the complexity of the sparse representation.

In the seakeeping application, the method outperforms quantile kriging at low and moderate budgets, and matches its best replication level at the largest budget, without selecting a replication level.

The present work also highlights several directions for future research.
From a theoretical perspective, a more detailed analysis of the surrogate objective and its impact on the location of the maximum \emph{a posteriori} estimate would further strengthen the foundations of the approach.
Similarly, establishing convergence guarantees for the sequential enrichment procedure remains an open problem.
Methodological extensions include an input-dependent asymmetric Laplace scale, simultaneous estimation of multiple quantile levels, and nonuniform integration weights targeting regions of interest, for example in quantile optimization.
These developments constitute promising directions for further advancing Bayesian quantile regression, particularly under limited computational and data-acquisition budgets.


\appendix


\section{Control of the lower-bound surrogate error} \label{app:control_of_the_lower_bound_surrogate_error}
For fixed inducing variables $\mathbf{u}$ and asymmetric Laplace scale parameter $\alpha > 0$, we write the values of the $\tau$-quantile function at the training inputs under the conditional prior as $\mu_n + \delta_n$, where $\mu_n := \mu ( \mathbf{x}_n \, ; \mathbf{u}, \mathbf{Z} )$ and $\boldsymbol{\delta} \sim \mathcal{N} ( \boldsymbol{0}, \mathbf{C}_N )$, with $( \mathbf{C}_N )_{ij} := \Sigma_z^2 ( \mathbf{x}_i, \mathbf{x}_j \, ; \mathbf{Z} )$.
The residuals $\delta_n$ are generally correlated, and the log-likelihood of the observations is
\begin{equation}
    L_N ( \boldsymbol{\delta} )
    :=
    N \log \! \left( \frac{ \tau ( 1 - \tau ) }{ \alpha } \right)
    - \frac{1}{\alpha} \sum_{n=1}^N \rho_\tau ( \mathrm{y}_n - \mu_n - \delta_n )
    .
    \label{eq:app_log_likelihood_residuals}
\end{equation}
The prior term cancels between the exact and surrogate objectives, so that
\begin{equation}
    \mathcal{J}_N ( \mathbf{u} ) - \widetilde{\mathcal{J}}_N ( \mathbf{u} )
    =
    \log \mathbb{E} \big[ e^{ L_N ( \boldsymbol{\delta} ) } \big]
    -
    \mathbb{E} \big[ L_N ( \boldsymbol{\delta} ) \big]
    ,
    \label{eq:app_objective_gap}
\end{equation}
where the expectations are taken under the joint conditional prior.

The check function \eqref{eq:check_function} is Lipschitz continuous with constant $L_\tau := \max ( \tau, 1 - \tau )$.
For any $\boldsymbol{\delta}, \boldsymbol{\delta}' \in \mathbb{R}^N$, we can therefore write $L_N ( \boldsymbol{\delta} ) - L_N ( \boldsymbol{\delta}' ) = \mathbf{a}^{\top} ( \boldsymbol{\delta} - \boldsymbol{\delta}' )$ for some vector $\mathbf{a}$, depending on $\boldsymbol{\delta}$ and $\boldsymbol{\delta}'$, with $\lvert a_n \rvert \leq L_\tau / \alpha$.
We write $\boldsymbol{\delta} = \mathbf{C}_N^{1/2} \mathbf{g}$, where $\mathbf{C}_N^{1/2}$ denotes the symmetric positive semidefinite square root of $\mathbf{C}_N$ and $\mathbf{g} \sim \mathcal{N} ( \boldsymbol{0}, \mathbf{I}_N )$.
The Cauchy--Schwarz inequality and the bound on $\mathbf{a}$ then give
\begin{equation}
    \big\lvert \mathbf{a}^{\top} \mathbf{C}_N^{1/2} ( \mathbf{g} - \mathbf{g}' ) \big\rvert^2
    \leq
    \mathbf{a}^{\top} \mathbf{C}_N \mathbf{a} \, \lVert \mathbf{g} - \mathbf{g}' \rVert_2^2
    \leq
    \frac{ L_\tau^2 }{ \alpha^2 } \sum_{i,j=1}^N \big\lvert ( \mathbf{C}_N )_{ij} \big\rvert \, \lVert \mathbf{g} - \mathbf{g}' \rVert_2^2
    .
    \label{eq:app_cauchy_schwarz}
\end{equation}
The function $h ( \mathbf{g} ) := L_N ( \mathbf{C}_N^{1/2} \mathbf{g} )$ is thus Lipschitz continuous with constant $c_N$, defined by
\begin{equation}
    c_N^2
    :=
    \frac{ L_\tau^2 }{ \alpha^2 } \sum_{i,j=1}^N \big\lvert ( \mathbf{C}_N )_{ij} \big\rvert
    .
    \label{eq:app_lipschitz_constant}
\end{equation}
The Gaussian concentration inequality~\cite[Theorem~5.5]{boucheron2013} states that $\log \mathbb{E} \big[ e^{ t ( h - \mathbb{E} [ h ] ) } \big] \leq t^2 c_N^2 / 2$ for all $t \in \mathbb{R}$.
Taking $t = 1$ and applying Jensen's inequality yields the nonasymptotic bound
\begin{equation}
    0
    \leq
    \frac{ \mathcal{J}_N ( \mathbf{u} ) - \widetilde{\mathcal{J}}_N ( \mathbf{u} ) }{ N }
    \leq
    \frac{ L_\tau^2 }{ 2 \alpha^2 N } \sum_{i,j=1}^N \big\lvert \Sigma_z^2 ( \mathbf{x}_i, \mathbf{x}_j \, ; \mathbf{Z} ) \big\rvert
    .
    \label{eq:joint_surrogate_gap_bound}
\end{equation}
This bound retains all conditional-prior covariances.
It holds for any positive semidefinite $\mathbf{C}_N$, including a singular one, as arises with replicated training inputs.
When $\mathbf{C}_N$ is diagonal, its right-hand side reduces to $L_\tau^2 / ( 2 \alpha^2 N ) \sum_{n=1}^N \sigma_n^2$, where $\sigma_n^2 := ( \mathbf{C}_N )_{nn}$.

For a fixed training dataset, $\lvert ( \mathbf{C}_N )_{ij} \rvert \leq \sigma_i \sigma_j$ implies that the objective gap vanishes as all conditional-prior variances at the training inputs vanish, provided that $\alpha$ remains bounded away from zero.
This connects the accuracy of the surrogate to the refinement of the inducing representation.
If $N$ also grows, a sufficient condition for the normalized gap to vanish is $N^{-1} \sum_{i,j=1}^N \lvert ( \mathbf{C}_N )_{ij} \rvert \to 0$, again with $\alpha$ bounded away from zero.
A vanishing average conditional-prior variance alone does not ensure this condition, which is not established here for the proposed sequential algorithm.


\section{Surrogate log-joint density and its Hessian} \label{app:analytical_expressions}
Here, we provide the expressions of the surrogate log-joint density $\widetilde{\mathcal{J}}_N$ and its Hessian.
Using the notation introduced in \cref{app:control_of_the_lower_bound_surrogate_error} and assuming $\sigma_n > 0$ for every $n$, the surrogate is given by
\begin{align}
    \widetilde{\mathcal{J}}_N ( \mathbf{u} \, ; \boldsymbol{\psi}, \alpha )        
    & =
    \mathbb{E}_{ \pi ( q_\tau \mid \mathbf{u}, \mathbf{Z}, \boldsymbol{\psi} ) } \big[ \log p ( \mathbf{y} \mid \mathbf{X}, q_\tau, \alpha ) \big]
    +
    \log \pi ( \mathbf{u} \mid \mathbf{Z}, \boldsymbol{\psi} )
    \\
    & =        
    - \frac{1}{\alpha} \sum_{n=1}^{N}
        \Bigg\{
            \left( \mathrm{y}_n - \mu_n \right)                
            \left[ \Phi \! \left( \frac{ \mathrm{y}_n - \mu_n } { \sigma_n  } \right) + \tau - 1 \right]
            + \sigma_n \, \varphi \! \left( \frac{ \mathrm{y}_n - \mu_n } {\sigma_n} \right)
            \nonumber
        \Bigg\}
    \\        
    & \hspace{0.42cm} - N \log \alpha 
    - \frac{1}{2} \mathbf{u}^{\top} \KzzHp^{-1} \mathbf{u} - \frac{1}{2} \log \left\lvert \KzzHp \right\rvert + \mathrm{C^{st}}
    \nonumber
    ,
\end{align}
where $\Phi$ and $\varphi$ denote the cumulative and probability density functions of the standard normal distribution, respectively.
The constant term $\mathrm{C^{st}}$ is independent of $\mathbf{u}$, $\boldsymbol{\psi}$, and $\alpha$.

The Hessian of the surrogate log-joint density $\widetilde{\mathcal{J}}_N$ with respect to $\mathbf{u}$ is given by
\begin{align}
    \nabla_{\mathbf{u}}^2 \, \widetilde{\mathcal{J}}_N ( \mathbf{u} \, ; \boldsymbol{\psi}, \alpha )    
    & =
    \nabla_{\mathbf{u}}^2 \, \mathbb{E}_{ \pi ( q_\tau \mid \mathbf{u}, \mathbf{Z}, \boldsymbol{\psi} ) } \big[ \log p ( \mathbf{y} \mid \mathbf{X}, q_\tau, \alpha ) \big]
    +
    \nabla_{\mathbf{u}}^2 \log \pi ( \mathbf{u} \mid \mathbf{Z}, \boldsymbol{\psi} )
    \\
    & =
    - \frac{1}{\alpha} \sum_{n=1}^{N}
        \Bigg\{            
            \KzzHp^{-1} K ( \mathbf{x}_n, \mathbf{Z} \, ; \boldsymbol{\psi} )^{\top}
            K ( \mathbf{x}_n, \mathbf{Z} \, ; \boldsymbol{\psi} ) \KzzHp^{-1}
            \nonumber 
            \\
            & \hspace{2.15cm} 
            \times
            \left[
                \frac{1}{ \sigma_n }
                \,
                \varphi \! \left(
                    \frac{ \mathrm{y}_n - \mu_n }{\sigma_n}
                \right)
            \right]
            \nonumber
        \Bigg\}        
    - \KzzHp^{-1}
    \nonumber
    .
\end{align}

\section*{Acknowledgments}
This work is supported by the LabCom MATritime funded by the Agence Nationale de la Recherche (Grant No.~ANR-22-LCV2-0010).
Numerical experiments presented in this paper were carried out using the PlaFRIM experimental testbed, supported by Inria, CNRS (LaBRI and IMB), Université de Bordeaux, Bordeaux INP, and Conseil Régional d'Aquitaine (see https://www.plafrim.fr).

\bibliographystyle{bstyle}
\bibliography{bfile}

\end{document}